# Generalizing Boolean Satisfiability II: Theory

**Heidi E. Dixon**                                                           dixon@otsys.com
*On Time Systems, Inc.*
*1850 Millrace, Suite 1*
*Eugene, OR 97403 USA*

**Matthew L. Ginsberg**                                                   ginsberg@otsys.com
*On Time Systems, Inc.*
*1850 Millrace, Suite 1*
*Eugene, OR 97403 USA*

**Eugene M. Luks**                                                       luks@cs.uoregon.edu
*Computer and Information Science*
*University of Oregon*
*Eugene, OR 97403 USA*

**Andrew J. Parkes**                                                   parkes@cirl.uoregon.edu
*CIRL*
*1269 University of Oregon*
*Eugene, OR 97403 USA*

## Abstract

This is the second of three planned papers describing ZAP, a satisfiability engine that substantially generalizes existing tools while retaining the performance characteristics of modern high performance solvers. The fundamental idea underlying ZAP is that many problems passed to such engines contain rich internal structure that is obscured by the Boolean representation used; our goal is to define a representation in which this structure is apparent and can easily be exploited to improve computational performance. This paper presents the theoretical basis for the ideas underlying ZAP, arguing that existing ideas in this area exploit a single, recurring structure in that multiple database axioms can be obtained by operating on a single axiom using a subgroup of the group of permutations on the literals in the problem. We argue that the group structure precisely captures the general structure at which earlier approaches hinted, and give numerous examples of its use. We go on to extend the Davis-Putnam-Logemann-Loveland inference procedure to this broader setting, and show that earlier computational improvements are either subsumed or left intact by the new method. The third paper in this series discusses ZAP's implementation and presents experimental performance results.

## 1. Introduction

This is the second of a planned series of three papers describing ZAP, a satisfiability engine that substantially generalizes existing tools while retaining the performance characteristics of existing high-performance solvers such as ZCHAFF (Moskewicz, Madigan, Zhao, Zhang, & Malik, 2001).[1] In the first paper (Dixon, Ginsberg, & Parkes, 2004b), to which we will refer as ZAP1, we discussed a variety of existing computational improvements to the

---

[1]. The first paper has appeared in JAIR; the third is currently available as a technical report (Dixon, Ginsberg, Hofer, Luks, & Parkes, 2004a) but has not yet been peer reviewed.

Davis-Putnam-Logemann-Loveland (DPLL) inference procedure, eventually producing the following table. The rows and columns are described on this page and the next.

| | efficiency of rep'n | proof length | resolution | propagation technique | learning method |
|---|---|---|---|---|---|
| **SAT** | — | EEE | — | watched literals | relevance |
| **cardinality** | exponential | P?E | not unique | watched literals | relevance |
| **pseudo-Boolean** | exponential | P?E | unique | watched literals | + strengthening |
| **symmetry** | — | EEE* | not believed in P | same as SAT | same as SAT |
| **QPROP** | exponential | ??? | in P using reasons | exp improvement | + first-order |

The rows of the table correspond to observations regarding existing representations used in satisfiability research, as reflected in the labels in the first column:[2]

1. **SAT** refers to conventional Boolean satisfiability work, representing information as conjunctions of disjunctions of literals (CNF).

2. **cardinality** refers to the use of "counting" clauses; if we think of a conventional disjunction of literals $\vee_i l_i$ as
$$\sum_i l_i \geq 1$$
then a cardinality clause is one of the form
$$\sum_i l_i \geq k$$
for a positive integer $k$.

3. **pseudo-Boolean** clauses extend cardinality clauses by allowing the literals in question to be weighted:
$$\sum_i w_i l_i \geq k$$
Each $w_i$ is a positive integer giving the weight to be assigned to the associated literal.

4. **symmetry** involves the introduction of techniques that are designed to explicitly exploit local or global symmetries in the problem being solved.

5. **QPROP** deals with universally quantified formulae where all of the quantifications are over finite domains of known size.

The columns in the table measure the performance of the various systems against a variety of metrics:

---

2. Please see the preceding paper ZAP1 (Dixon et al., 2004b) for a fuller explanation and for a relatively comprehensive list of references where the earlier work is discussed.

1. **Efficiency of representation** measures the extent to which a single axiom in a proposed framework can replace many in CNF. For cardinality, pseudo-Boolean and quantified languages, it is possible that exponential savings are achieved. We argued that such savings were possible but relatively unlikely for cardinality and pseudo-Boolean encodings but were relatively likely for QPROP.

2. **Proof length** gives the minimum proof length for the representation on three classes of problems: the pigeonhole problem, parity problems due to Tseitin (1970) and clique coloring problems (Pudlak, 1997). An E indicates exponential proof length; P indicates polynomial length. While symmetry-exploitation techniques can provide polynomial-length proofs in certain instances, the method is so brittle against changes in the axiomatization that we do not regard this as a polynomial approach in general.

3. **Resolution** indicates the extent to which resolution can be lifted to a broader setting. This is straightforward in the pseudo-Boolean case; cardinality clauses have the problem that the most natural resolvent of two cardinality clauses may not be a cardinality clause, and there may be many cardinality clauses that could be derived as a result. Systems that exploit local symmetries must search for such symmetries at each inference step, a problem that is not believed to be in P. Provided that reasons are maintained, inference remains well defined for quantified axioms, requiring only the introduction of a linear complexity unification step.

4. **Propagation technique** describes the techniques used to draw conclusions from an existing partial assignment of values to variables. For all of the systems except QPROP, Zhang and Stickel's watched literals idea (Moskewicz et al., 2001; Zhang & Stickel, 2000) is the most efficient mechanism known. This approach cannot be lifted to QPROP, but a somewhat simpler method can be lifted and average-case exponential savings obtained as a result (Ginsberg & Parkes, 2000).

5. **Learning method** gives the technique typically used to save conclusions as the inference proceeds. In general, relevance-bounded learning (Bayardo & Miranker, 1996; Bayardo & Schrag, 1997; Ginsberg, 1993) is the most effective technique known here. It can be augmented with strengthening (Guignard & Spielberg, 1981; Savelsbergh, 1994) in the pseudo-Boolean case and with first-order reasoning if quantified formulae are present.

Our goal in the current paper is to add a single line to the above table:

|  | efficiency of rep'n | proof length | resolution | propagation technique | learning method |
|---|---|---|---|---|---|
| SAT | — | EEE | — | watched literals | relevance |
| cardinality | exponential | P?E | not unique | watched literals | relevance |
| pseudo-Boolean | exponential | P?E | unique | watched literals | + strengthening |
| symmetry | — | EEE* | not believed in P | same as SAT | same as SAT |
| QPROP | exponential | ??? | in P using reasons | exp improvement | + first-order |
| ZAP | exponential | PPP | in P using reasons | watched literals, exp improvement | + first-order + parity + others |

ZAP is the approach to inference that is the focus of this series of papers. The basic idea in ZAP is that in realistic problems, many Boolean clauses are "versions" of a single clause. We will make this notion precise shortly; at this point, one might think of all of the instances of a quantified clause as being versions of any particular ground instance. The versions, it will turn out, correspond to permutations on the set of literals in the problem.

As an example, suppose that we are tying to prove that it is impossible to put $n+1$ pigeons into $n$ holes if each pigeon is to get its own hole. A Boolean axiomatization of this problem will include the axioms

$$\begin{array}{cccc}
\neg p_{11} \vee \neg p_{21} & \neg p_{12} \vee \neg p_{22} & \cdots & \neg p_{1n} \vee \neg p_{2n} \\
\neg p_{11} \vee \neg p_{31} & \neg p_{12} \vee \neg p_{32} & \cdots & \neg p_{1n} \vee \neg p_{3n} \\
\vdots & \vdots & & \vdots \\
\neg p_{11} \vee \neg p_{n+1,1} & \neg p_{12} \vee \neg p_{n+1,2} & \cdots & \neg p_{1n} \vee \neg p_{n+1,n} \\
\neg p_{21} \vee \neg p_{31} & \neg p_{22} \vee \neg p_{32} & \cdots & \neg p_{2n} \vee \neg p_{3n} \\
\vdots & \vdots & & \vdots \\
\neg p_{21} \vee \neg p_{n+1,1} & \neg p_{22} \vee \neg p_{n+1,2} & \cdots & \neg p_{2n} \vee \neg p_{n+1,n} \\
\vdots & \vdots & & \vdots \\
\neg p_{n1} \vee \neg p_{n+1,1} & \neg p_{n2} \vee \neg p_{n+1,2} & \cdots & \neg p_{nn} \vee \neg p_{n+1,n}
\end{array}$$

where $p_{ij}$ says that pigeon $i$ is in hole $j$. Thus the first clause above says that pigeon one and pigeon two cannot both be in hole one. The second clause in the first column says that pigeon one and pigeon three cannot both be in hole one. The second column refers to hole two, and so on. It is fairly clear that all of these axioms can be reconstructed from the first by interchanging the pigeons and the holes, and it is this intuition that ZAP attempts to capture.

What makes this approach interesting is the fact that instead of reasoning with a large set of clauses, it becomes possible to reason with a single clause instance and a set of permutations. As we will see, the sets of permutations that occur naturally are highly structured sets called *groups*, and exploiting this structure can lead to significant efficiency gains in both representation and reasoning.

Some further comments on the above table:

- Unlike cardinality and pseudo-Boolean methods, which seem unlikely to achieve exponential reductions in problem size in practice, and QPROP, which seems likely to achieve such reductions, ZAP is *guaranteed*, when the requisite structure is present, to replace a set of $n$ axioms with a single axiom of size at most $v \log(n)$, where $v$ is the number of variables in the problem (Proposition 4.8).

- The fundamental inference step in ZAP is in NP with respect to the ZAP representation, and therefore has complexity no worse than exponential in the representation size (i.e., polynomial in the number of Boolean axioms being resolved). In practice, the average case complexity appears to be low-order polynomial in the size of the ZAP representation (i.e., polynomial in the logarithm of the number of Boolean axioms being resolved) (Dixon et al., 2004a).

- ZAP obtains the savings attributable to subsearch in the QPROP case while casting them in a general setting that is equivalent to watched literals in the Boolean case. This particular observation is dependent on a variety of results from computational group theory and is discussed in the third paper in this series (Dixon et al., 2004a).

- In addition to learning the Boolean consequences of resolution, ZAP continues to support relevance-bounded learning schemes while also allowing the derivation of first-order consequences, conclusions based on parity arguments, and combinations thereof.

In order to deliver on these claims, we begin in Section 2 by summarizing both the DPLL algorithm and the modifications that embody recent progress, casting DPLL into the precise form that is needed in ZAP and that seems to best capture the architecture of modern systems such as zCHAFF. Section 3 is also a summary of ideas that are not new with us, providing an introduction to some ideas from group theory.

In Section 4, we describe the key insight underlying ZAP. As mentioned above, the structure exploited in earlier examples corresponds to the existence of particular groups of permutations of the literals in the problem. We call the combination of a clause and such a permutation group an *augmented clause*, and the **efficiency of representation** column of our table corresponds to the observation that augmented clauses can use group structure to improve the efficiency of their encoding.

Section 5 (**resolution**) describes resolution in this broader setting, and Section 6 (**proof length**) presents a variety of examples of these ideas at work, showing that the pigeonhole problem, clique-coloring problems, and Tseitin's parity examples all admit short proofs in the new framework. Section 7 (**learning method**) recasts the DPLL algorithm in the new terms and discusses the continued applicability of relevance in our setting. Concluding remarks are contained in Section 8. Implementation details, including a discussion of **propagation techniques**, are deferred until the third of this series of papers (Dixon et al., 2004a). This third paper will also include a presentation of performance details; at this point, we note merely that ZAP does indeed exhibit polynomial performance on the natural encodings of pigeonhole, parity and clique-coloring problems. This is in sharp contrast with other methods, where theoretical best-case performance (let alone experimental average-case performance) is known to be exponential on these problems classes.

## 2. Boolean Satisfiability Engines

In ZAP1, we presented descriptions of the standard Davis-Putnam-Logemann-Loveland (DPLL) Boolean satisfiability algorithm and described informally the extensions to DPLL that deal with learning. Our goal in this paper and the next is to describe an implementation of our theoretical ideas. We therefore begin here by being more precise about DPLL and its extension to relevance-bounded learning, or RBL. We present some general definitions that we will need throughout this paper, and then give a description of the DPLL algorithm in a learning/reason-maintenance setting. We prove that an implementation of these ideas can retain the soundness and completeness of DPLL while using an amount of memory that grows polynomially with problem size. Although this result has been generally accepted since 1-relevance learning ("dynamic backtracking," Ginsberg, 1993) was generalized by Bayardo, Miranker and Schrag (Bayardo & Miranker, 1996; Bayardo & Schrag, 1997), we know of no previous proof that RBL has the stated properties.

**Definition 2.1** *A* partial assignment *is an ordered sequence*

$$\langle l_1, \ldots, l_n \rangle$$

*of distinct and consistent literals.*

**Definition 2.2** *Let $\vee_i l_i$ be a clause, which we will denote by $c$, and suppose that $P$ is a partial assignment. We will say that the* possible value of $c$ under $P$ *is given by*

$$\text{poss}(c, P) = |\{l \in c | \neg l \notin P\}| - 1$$

*If no ambiguity is possible, we will write simply $\text{poss}(c)$ instead of $\text{poss}(c, P)$. In other words, $\text{poss}(c)$ is the number of literals that are either already satisfied or not valued by $P$, reduced by one (since true clauses require at least one true literal).*

*If $S$ is a set of clauses, we will write $\text{poss}_n(S, P)$ for the subset of $c \in S$ for which $\text{poss}(c, P) \leq n$.*

*In a similar way, we will define $\text{curr}(c, P)$ to be*

$$\text{curr}(c, P) = |\{l \in c \cap P\}| - 1$$

*We will write $\text{curr}_n(S, P)$ for the subset of $c \in S$ for which $\text{curr}(c, P) \leq n$.*

Informally, if $\text{poss}(c, P) = 0$, that means that any partial assignment extending $P$ can make at most one literal in $c$ true; there is no room for any "extra" literals to be true. This might be because all of the literals in $c$ are assigned values by $P$ and only one such literal is true; it might be because there is a single unvalued literal and all of the other literals are false. If $\text{poss}(c, P) < 0$, it means that the given partial assignment cannot be extended in a way that will cause $c$ to be satisfied. Thus we see that $\text{poss}_{-1}(S, P)$ is the set of clauses in $S$ that are falsified by $P$. Since curr gives the "current" excess in the number of satisfied literals (as opposed to poss, which gives the possible excess), the set $\text{poss}_0(S, P) \cap \text{curr}_{-1}(S, P)$ is the set of clauses that are not currently satisfied and have at most one unvalued literal. These are generally referred to as *unit* clauses.

We note in passing that Definition 2.2 can easily be extended to deal with pseudo-Boolean instead of Boolean clauses, although that extension will not be our focus here.

**Definition 2.3** *An* annotated partial assignment *is an ordered sequence*

$$\langle (l_1, c_1), \ldots, (l_n, c_n) \rangle$$

*of distinct and consistent literals and clauses, where $c_i$ is the* reason *for literal $l_i$ and either $c_i = $* true *(indicating that $l_i$ was a branch point) or $c_i$ is a clause such that:*

1. *$l_i$ is a literal in $c_i$, and*

2. $\mathtt{poss}(c_i, \langle l_1, \ldots, l_{i-1} \rangle) = 0$

*An annotated partial assignment will be called* sound *with respect to a set of clauses $C$ if $C \models c_i$ for each reason $c_i$.*

The reasons have the property that after the literals $l_1, \ldots, l_{i-1}$ are all included in the partial assignment, it is possible to conclude $l_i$ directly from $c_i$, since there is no other way for $c_i$ to be satisfied.

**Definition 2.4** *Let $c_1$ and $c_2$ be clauses, each a set of literals to be disjoined. We will say that $c_1$ and $c_2$* resolve *if there is a unique literal $l$ such that $l \in c_1$ and $\neg l \in c_2$. If $c_1$ and $c_2$ resolve, their* resolvent, *to be denoted* $\mathtt{resolve}(c_1, c_2)$, *is the disjunction of the literals in the set $c_1 \cup c_2 - \{l, \neg l\}$.*

*If $r_1$ and $r_2$ are reasons, the result of resolving $r_1$ and $r_2$ is defined to be:*

$$\mathtt{resolve}(r_1, r_2) = \begin{cases} r_2, & \text{if } r_1 = \mathtt{true}; \\ r_1, & \text{if } r_2 = \mathtt{true}; \\ \text{the conventional resolvent of } r_1 \text{ and } r_2, & \text{otherwise.} \end{cases}$$

**Definition 2.5** *Let $C$ be a set of clauses and $P$ a partial assignment. A* nogood *for $P$ is any clause $c$ that is entailed by $C$ but falsified by $P$. A* nogood *is any clause $c$ that is entailed by $C$.*

We are now in a position to present one of the basic building blocks of DPLL or RBL, the *unit propagation* procedure. This computes the "obvious" consequences of a partial assignment:

**Procedure 2.6 (Unit propagation)** *To compute* UNIT-PROPAGATE$(C, P)$ *for a set $C$ of clauses and an annotated partial assignment $P$:*

```
1   while poss₀(C, P) ∩ curr₋₁(C, P) ≠ ∅
2       do if poss₋₁(C, P) ≠ ∅
3           then c ← an element of poss₋₁(C, P)
4                lᵢ ← the literal in c with the highest index in P
5                return ⟨true, resolve(c, cᵢ)⟩
6           else c ← an element of poss₀(C, P) ∩ curr₋₁(C, P)
7                l ← the literal in c unassigned by P
8                P ← ⟨P, (l, c)⟩
9   return ⟨false, P⟩
```

The above procedure returns a pair of values. The first indicates whether a contradiction has been found. If so, the second value is the reason for the failure, a consequence of the clausal database $C$ that is falsified by the partial assignment $P$ (i.e., a nogood). If no contradiction is found, the second value is a suitably extended partial assignment. Procedure 2.6 has also been modified to work with annotated partial assignments, and to annotate the new choices that are made when $P$ is extended.

**Proposition 2.7** *Suppose that $C$ is a Boolean satisfiability problem, and $P$ is a sound annotated partial assignment. Then:*

1. *If* unit-propagate$(P) = \langle \text{false}, P' \rangle$, *then $P'$ is a sound extension of $P$, and*

2. *If* unit-propagate$(P) = \langle \text{true}, c \rangle$, *then $c$ is a nogood for $P$.*

**Proof.** In the interests of maintaining the continuity of the exposition, most proofs (including this one!) have been deferred to the appendix. Proofs or proof sketches will appear in the main text only when they further the exposition in some way. □

We can now describe relevance-bounded learning:

**Procedure 2.8 (Relevance-bounded reasoning, RBL)** *Let $C$ be a* SAT *problem, and $D$ a set of learned nogoods, so that $C \models D$. Let $P$ be an annotated partial assignment, and $k$ a fixed relevance bound. To compute* RBL$(C, D, P)$:

```
1   ⟨x, y⟩ ← Unit-Propagate(C ∪ D, P)
2   if x = true
3       then c ← y
4           if c is empty
5               then return FAILURE
6               else  remove successive elements from P so that c is satisfiable and
                         poss(c, P) ≤ k
7                    D ← {c} ∪ poss_k(D, P)
8                    return RBL(C, D, P)
9       else  P ← y
10          if P is a solution to C
11              then return P
12              else  l ← a literal not assigned a value by P
13                   return RBL(C, D, ⟨P, (l, true)⟩)
```

This procedure is fairly different from the original description of DPLL, so let us go through it.

In general, variables are assigned values either via branching on line 13, or unit propagation on lines 1 and 9. If unit propagation terminates without reaching a contradiction or finding a solution (so that $x$ is false on line 2 and the test on line 10 fails as well), then a branch variable is selected and assigned a value, and the procedure recurs.

If the unit propagation procedure "fails" and returns $\langle \text{true}, c \rangle$ for a new nogood $c$, the new clause is learned by adding it to $D$, and the search backtracks at least to the point where $c$ is satisfiable.[3] Any learned clauses that have become irrelevant (in that their poss value exceeds the irrelevance cutoff $k$) are removed. Note that we only remove *learned* irrelevant nogoods; we obviously cannot remove clauses that were part of the original problem specification. It is for this reason that the sets $C$ and $D$ (of original and learned clauses respectively) are maintained separately.

Procedure 2.8 can fail only if a contradiction (an empty clause $c$) is derived. In all other cases, progress is made by augmenting the set of clauses to include at least one new nogood that eliminates the current partial assignment. Instead of resetting the branch literal $l$ to take the opposite value as in Davis et.al.'s original description of their algorithm, a new clause is learned and added to the problem. This new clause causes either $l$ or some previous variable to take a new value.

The above description is ambiguous about a variety of points. We do not specify how far to backtrack on line 6, or the branch literal to be chosen on line 12. We will not be concerned with these choices; ZAP takes the same approach that ZCHAFF does and the implementation is straightforward.

**Theorem 2.9** RBL *is sound and complete in that* $\text{RBL}(C, \emptyset, \langle \rangle)$ *will always return a satisfying assignment if $C$ is satisfiable and will always report failure if $C$ is unsatisfiable.* RBL *also uses an amount of memory polynomial in the size of $C$ (although exponential in the relevance bound $k$).*

As discussed at some length in ZAP1, other authors have focused on extending the language of Boolean satisfiability in ways that preserve the efficiency of RBL. These extensions include the ability to deal with quantification (Ginsberg & Parkes, 2000), pseudo-Boolean or cardinality clauses (Barth, 1995, 1996; Chandru & Hooker, 1999; Dixon & Ginsberg, 2000; Hooker, 1988; Nemhauser & Wolsey, 1988), or parity clauses (Baumgartner & Massacci, 2000; Li, 2000).

## 3. Some Concepts from Group Theory

Relevance-bounded learning is only a part of the background that we will need to describe ZAP; we also need some basic ideas from group theory. There are many excellent references available on this topic (Rotman, 1994, and others), and we can only give a brief account here. Our goal is not to present a terse sequence of definitions and to then hollowly claim that this paper is self-contained; we would rather provide insight regarding the goals and underlying philosophy of group theory generally. We will face a similar problem in the final paper in this series, where we will draw heavily on results from computational group theory and will, once again, present a compact and hopefully helpful overview of a broad area of mathematics.

**Definition 3.1** *A* group *is a set $S$ equipped with an associative binary operator $\circ$. The operator $\circ$ has an identity $1$, with $1 \circ x = x \circ 1 = x$ for all $x \in S$, and an inverse, so that for every $x \in S$ there is an $x^{-1}$ such that $x \circ x^{-1} = x^{-1} \circ x = 1$.*

---

3. As we remarked in ZAP1, some systems such as ZCHAFF (Moskewicz et al., 2001) backtrack further to the point where $c$ is unit.

In other words, $\circ$ is a function $\circ : S \times S \rightarrow S$; since $\circ$ is associative, we always have

$$(x \circ y) \circ z = x \circ (y \circ z)$$

The group operator $\circ$ is not required to be commutative; if it is commutative, the group is called *Abelian*.

Typical examples of groups include $\mathbb{Z}$, the group of integers with the operation being addition. Similarly, $\mathbb{Q}$ and $\mathbb{R}$ are the groups of rationals or reals under addition. For multiplication, zero needs to be excluded, since it has no inverse, and we get the groups $\mathbb{Q}^*$ and $\mathbb{R}^*$. Note that $\mathbb{Z}^*$ is not a group, since $1/n$ is not an integer for most integers $n$.

Other common groups include $\mathbb{Z}_n$ for any positive integer $n$; this is the group of integers mod $n$, where the group operation is addition mod $n$. For a prime $p$, the set $\mathbb{Z}_p^*$ of nonzero integers mod $p$ does have a multiplicative inverse, so that $\mathbb{Z}_p^*$ is a group under multiplication. The group $\mathbb{Z}_1$ contains the single element 0 and is the *trivial* group. This group is often denoted by **1**.

All of the groups we have described thus far are Abelian, but non-Abelian groups are not hard to come by. As an example, the set of all $n \times n$ matrices with real entries and nonzero determinants is a group under multiplication, since every matrix with a nonzero determinant has an inverse. This group is called the *general linear group* and is denoted $GL(n)$.

Of particular interest to us will be the so-called *permutation groups*:

**Definition 3.2** *Let $T$ be a set. A* permutation *of $T$ is a bijection $\omega : T \rightarrow T$ from $T$ to itself.*

**Proposition 3.3** *Let $T$ be a set. Then the set of permutations of $T$ is a group under the composition operator.*

**Proof.** This is simply a matter of validating the definition. Functional composition is well known to be associative (although not necessarily commutative), and the identity function from $T$ to itself is the identity for the composition operator. Since each permutation is a bijection, permutations can be inverted to give the inverse operator for the group. □

The group of permutations on $T$ is called the *symmetry group of $T$*, and is typically denoted $\text{Sym}(T)$. We will take the view that the composition $f \circ g$ acts with $f$ first and $g$ second, so that $(f \circ g)(x) = g(f(x))$ for any $x \in T$. (Note the variable order.)

Because permutation groups will be of such interest to us, it is worthwhile to introduce some additional notation for dealing with them in the case where $T \subseteq \mathbb{Z}$ is a subset of the integers. In the special case where $T = \{1, \ldots, n\}$, we will often abbreviate $\text{Sym}(T)$ to either $\text{Sym}(n)$ or simply $S_n$.

Of course, we can get groups of permutations without including *every* permutation on a particular set; the 2-element set consisting of the identity permutation and the permutation that swaps two specific elements of $T$ is closed under inversion and composition and is therefore a group as well. In general, we have:

**Definition 3.4** *Let $(G, \circ)$ be a group. Then a subset $H \subseteq G$ is called a* subgroup *if $(H, \circ)$ is a group. This is denoted $H \leq G$. If the inclusion is proper, we write $H < G$.*

A subgroup of a group $G$ is any subset of $G$ that includes the identity and that is closed under composition and inversion.

If $G$ is a finite group, closure under composition suffices. To understand why, suppose that we have some subset $H \subseteq G$ that is closed under composition, and that $x \in H$. Now $x^2 \in H$, and $x^3 \in H$, and so on. Since $G$ is finite, we must eventually have $x^m = x^n$ for some integers $m$ and $n$, where we assume $m > n$. It follows that $x^{m-n} = 1$ so $x^{m-n-1} = x^{-1}$, so $H$ is closed under inversion and therefore a subgroup.

**Proposition 3.5** *Let $G$ be a group, and suppose that $H_1 \leq G$ and $H_2 \leq G$ are subgroups. Then $H_1 \cap H_2 \leq G$ is a subgroup of $G$ as well.* ▫

This should be clear, since $H_1 \cap H_2$ must be closed under inversion and composition if each of the component groups is.

**Definition 3.6** *Let $G$ be a group, and $S \subseteq G$ a subset. Then there is a unique smallest subgroup of $G$ containing $S$, which is denoted $\langle S \rangle$ and called the subgroup of $G$ generated by $S$. The* order *of an element $g \in G$ is defined to be $|\langle g \rangle|$.*

The generated subgroup is unique because it can be formed by taking the intersection of all subgroups containing $S$. This intersection is itself a subgroup by virtue of Proposition 3.5. If $S = \emptyset$ or $S = \{1\}$, the trivial subgroup is generated, consisting of only the identity element of $G$. Thus the order of the identity element is one. For any element $g$, the order of $g$ is the least nonzero exponent $m$ for which $g^m = 1$.

We have already remarked that the two-element set $\{1, (ab)\}$ is a group, where 1 represents the trivial permutation and $(ab)$ is the permutation that swaps $a$ and $b$. It is easy to see that $\{1, (ab)\}$ is the group generated by $(ab)$. The order of $(ab)$ is two.

In a similar way, if $(abcd)$ is the permutation that maps $a$ to $b$, $b$ to $c$, $c$ to $d$ and $d$ back to $a$, then the subgroup generated by $(abcd)$ is

$$\{1, (abcd), (ac) \circ (bd), (adcb)\}$$

The third element simultaneously swaps $a$ and $c$, and swaps $b$ and $d$. The order of the permutation $(abcd)$ is four, and $(abcd)$ is called a *4-cycle*. Both this subgroup and the previous subgroup generated by $(ab)$ are Abelian, although the full permutation group $\mathrm{Sym}(L)$ is not Abelian if $|L| > 2$. It is not hard to see that $\langle \rho \rangle$ is Abelian for any specific permutation $\rho$.

Slightly more interesting is the group generated by $(abcd)$ together with $(ac)$. This group has eight elements:

$$\{1, (abcd), (ac) \circ (bd), (adcb), (ac), (ad) \circ (bc), (bd), (ab) \circ (cd)\} \qquad (1)$$

The first four permutations don't "use" $(ac)$ and the second four do. Since $(abcd) \circ (ac) \neq (ac) \circ (abcd)$, this group is not Abelian.

It is not hard to see that the group (1) is in fact the group of rigid motions of a square whose vertices are labeled $a$, $b$, $c$ and $d$. The first permutation $(abcd)$ corresponds to a rotation of the square by $90°$ and the second $(ac)$, to a flip around the $b$-$d$ diagonal. The first four permutations in (1) simply rotate the square, while the second four use the flip as well; the group is not Abelian because a flip followed by a 90° clockwise rotation is different than the rotation followed by the flip. In a similar way, the six basic twists of Rubik's cube generate a permutation group of size approximately $4.3 \times 10^{19}$, giving all accessible permutations of the faces.

In general, suppose that $f$ is a permutation on a set $S$ and $x \in S$. We can obviously consider the image of $x$ under $f$. Rather than denote this image by $f(x)$ as usual, it is customary to denote it by $x^f$. The reason for this "inline" notation is that we now have

$$x^{fg} = (x^f)^g$$

which seems natural, as opposed to the unnatural

$$(fg)(x) = g(f(x))$$

as mentioned previously. We have dropped the explicit composition operator ∘ here.

Continuing, we can form the set of images $x^f$ where $f$ varies over all elements of some permutation group $G$. This is called the *orbit* of $x$ under $G$:

**Definition 3.7** *Let $G \leq \mathrm{Sym}(T)$ be a permutation group. Then for any $x \in T$, the* orbit *of $x$ in $G$, to be denoted by $x^G$, is given by $x^G = \{x^g | g \in G\}$.*

Returning to the case of permutations on integers, suppose that $n$ is an integer and $\omega$ a permutation. We can consider $\langle \omega \rangle$, the group of permutations generated by $\omega$, which is the set of powers of $\omega$ until we eventually have $\omega^m = 1$ for $m$ the order of $\omega$. The orbit of $n$ under $\langle \omega \rangle$ is the set $n^{\langle \omega \rangle} = \{n, n^\omega, n^{\omega^2}, \ldots, n^{\omega^{m-1}}\}$.

Now suppose that $n'$ is some other integer that appears in this sequence, say $n' = n^{\omega^k}$. Now $n'^\omega = (n^{\omega^k})^\omega = n^{\omega^{k+1}}$, so that the images of $n'$ can be "read off" from the sequence of images of $n$. It therefore makes sense to write this "piece" of the permutation as (for example)

$$(1, 3, 4) \tag{2}$$

indicating that 1 is mapped to 3, that 3 is mapped to 4, and that 4 is mapped back to 1.

Of course, the 3-cycle (2) doesn't tell us what happens to integers that are not in $n^{\langle \omega \rangle}$; for them, we need another cycle as in (2). So if the permutation $\omega$ swaps 2 and 5 in addition to mapping 1 to 3 and so on, we might write

$$\omega = (1, 3, 4)(2, 5) \tag{3}$$

If 6 is not moved by $\omega$ (so that $6^\omega = 6$), we could write

$$\omega = (1, 3, 4)(2, 5)(6) \tag{4}$$

In general, we will not mention variables that are fixed by the permutation, preferring (3) to the longer (4). We can often omit the commas within the cycles, so that we will continue to abbreviate $(a, b, c)$ as simply $(abc)$. If we need to indicate explicitly that two cycles are part of a single permutation, we will introduce an extra set of parentheses, perhaps rewriting (3) as

$$\omega = ((1, 3, 4)(2, 5))$$

Every permutation can be written as a product of disjoint cycles in this way.

Finally, in composing permutations written in this fashion, we will either drop the ∘ or replace it with ·, so that we have, for example,

$$(abc) \cdot (abd) = (ad)(bc)$$

The point $a$ is moved to $b$ by the first cycle and then to $d$ by the second. The point $b$ is moved to $c$ by the first cycle and then not changed by the second; $c$ is moved to $a$ and then on to $b$. Finally, $d$ is not moved by the first cycle but is moved to $a$ by the second.

Two other notions that we will need are that of closure and stabilizer:

**Definition 3.8** *Let $G \leq \text{Sym}(T)$, and $S \subseteq T$. By the G-closure of $S$, to be denoted $S^G$, we will mean the set*
$$S^G = \{s^g | s \in S \text{ and } g \in G\}$$

**Definition 3.9** *Given a group $G \leq \text{Sym}(T)$ and $L \subseteq T$, the* pointwise stabilizer *of $L$, denoted $G_L$, is the subgroup of all $g \in G$ such that $l^g = l$ for every $l \in L$. The* set stabilizer *of $L$, denoted $G_{\{L\}}$, is that subgroup of all $g \in G$ such that $L^g = L$.*

As an example, consider the group $G$ generated by the permutation $\omega = (1, 3, 4)(2, 5)$ that we considered above. Since $\omega^2 = (1, 4, 3)$ and $\omega^3 = (2, 5)$, it is not too hard to see that $G = \langle (1, 4, 3), (2, 5) \rangle$ is the group generated by the 3-cycle $(1, 4, 3)$ and the 2-cycle $(2, 5)$. The subgroup of $G$ that point stabilizes the set $\{2\}$ is thus $G_2 = \langle (1, 4, 3) \rangle$, and $G_{2,5}$ is identical. The subgroup of $G$ that *set* stabilizes $\{2, 5\}$ is $G_{\{2,5\}} = G$, however, since every permutation in $G$ leaves the set $\{2, 5\}$ intact.

## 4. Axiom Structure as a Group

While we will need the details of Procedures 2.8 and 2.6 in order to implement our ideas, the procedures themselves inherit certain weaknesses of DPLL as originally described. Two weaknesses that we hope to address are:

1. The appearance of $\text{poss}_0(C, P) \cap \text{curr}_{-1}(C, P)$ in the inner unit propagation loop requires an examination of a significant subset of the clausal database at each inference step, and

2. Both DPLL and RBL are fundamentally resolution-based methods; there are known problem classes that are exponentially difficult for resolution-based methods but which are easy if the language in use is extended to include either cardinality or parity clauses.

### 4.1 Examples of Structure

Let us begin by examining examples where specialized techniques can help to address these difficulties.

### 4.1.1 Subsearch

As we have discussed elsewhere (Dixon et al., 2004b; Ginsberg & Parkes, 2000), the set of axioms that need to be investigated in the DPLL inner loop often has structure that can be exploited to speed the examination process. If a ground axiomatization is replaced with a lifted one, the search for axioms with specific syntactic properties is NP-complete in the number of variables in the lifted axiom, and is called *subsearch* for that reason.

In many cases, search techniques can be applied to the subsearch problem. As an example, suppose that we are looking for instances of the lifted axiom

$$a(x,y) \vee b(y,z) \vee c(x,z) \tag{5}$$

that are unit, so that $\texttt{poss}(i,P) = 0$ and $\texttt{curr}(i,P) = -1$ for some such instance $i$ and a unit propagation is possible as a result.

Our notation here is that of QPROP. There is an implicit universal quantification over $x$, $y$ and $z$, and each quantification is over a domain of finite size. We assume that all of the domains are of size $d$, so (5) corresponds to $d^3$ ground axioms. If $a(x,y)$ is true for all $x$ and $y$ (which we can surely conclude in time $O(d^2)$), then we can conclude without further work that (5) has no unit instances, since every instance of (5) is already satisfied. If $a(x,y)$ is true except for a single $(x,y)$ pair, then we need only examine the $d$ possible values of $z$ for unit instances, reducing our total work from $d^3$ to $d^2 + d$.

It will be useful in what follows to make this example still more specific, so let us assume that $x$, $y$ and $z$ are all chosen from a two element domain $\{A, B\}$. The single lifted axiom (5) now corresponds to the set of ground instances:

$$a(A,A) \vee b(A,A) \vee c(A,A)$$
$$a(A,A) \vee b(A,B) \vee c(A,B)$$
$$a(A,B) \vee b(B,A) \vee c(A,A)$$
$$a(A,B) \vee b(B,B) \vee c(A,B)$$
$$a(B,A) \vee b(A,A) \vee c(B,A)$$
$$a(B,A) \vee b(A,B) \vee c(B,B)$$
$$a(B,B) \vee b(B,A) \vee c(B,A)$$
$$a(B,B) \vee b(B,B) \vee c(B,B)$$

If we introduce ground literals $l_1, l_2, l_3, l_4$ for the instances of $a(x,y)$ and so on, we get:

$$\begin{aligned} l_1 &\vee l_5 \vee l_9 \\ l_1 &\vee l_6 \vee l_{10} \\ l_2 &\vee l_7 \vee l_9 \\ l_2 &\vee l_8 \vee l_{10} \\ l_3 &\vee l_5 \vee l_{11} \\ l_3 &\vee l_6 \vee l_{12} \\ l_4 &\vee l_7 \vee l_{11} \\ l_4 &\vee l_8 \vee l_{12} \end{aligned} \tag{6}$$

at which point the structure implicit in (5) has been obscured. We will return to the details of this example shortly.

### 4.1.2 Cardinality

Structure is also present in the sets of axioms used to encode the pigeonhole problem, which is known to be exponentially difficult for any resolution-based method (Haken, 1985). As shown by a variety of authors (Cook, Coullard, & Turan, 1987; Dixon & Ginsberg, 2000), the pigeonhole problem can be solved in polynomial time if we extend our representation to include cardinality axioms such as

$$x_1 + \cdots + x_m \geq k \qquad (7)$$

As shown in ZAP1, the single axiom (7) is equivalent to $\binom{m}{k-1}$ conventional disjunctions.

As in Section 4.1.1, we will consider this example in detail. Suppose that we have the clause

$$x_1 + x_2 + x_3 + x_4 + x_5 \geq 3 \qquad (8)$$

saying that at least 3 of the $x_i$'s are true. This is equivalent to

$$\begin{array}{ll} x_1 \vee x_2 \vee x_3 & x_1 \vee x_4 \vee x_5 \\ x_1 \vee x_2 \vee x_4 & x_2 \vee x_3 \vee x_4 \\ x_1 \vee x_2 \vee x_5 & x_2 \vee x_3 \vee x_5 \\ x_1 \vee x_3 \vee x_4 & x_2 \vee x_4 \vee x_5 \\ x_1 \vee x_3 \vee x_5 & x_3 \vee x_4 \vee x_5 \end{array} \qquad (9)$$

### 4.1.3 Parity Clauses

Finally, we consider clauses that are most naturally expressed using modular arithmetic or exclusive or's, such as

$$x_1 + \cdots + x_k \equiv 0 \pmod{2} \qquad (10)$$

or

$$x_1 + \cdots + x_k \equiv 1 \pmod{2} \qquad (11)$$

The parity of the sum of the $x_i$'s is specified as even in (10) or as odd in (11).

It is well known that axiom sets consisting of parity clauses in isolation can be solved in polynomial time using Gaussian elimination, but there are examples that are exponentially difficult for resolution-based methods (Tseitin, 1970). As in the other examples we have discussed, single axioms such as (11) reveal structure that a straightforward Boolean axiomatization obscures. In this case, the single axiom (11) with $k = 3$ is equivalent to:

$$\begin{array}{l} x_1 \vee x_2 \vee x_3 \\ x_1 \vee \neg x_2 \vee \neg x_3 \\ \neg x_1 \vee x_2 \vee \neg x_3 \\ \neg x_1 \vee \neg x_2 \vee x_3 \end{array} \qquad (12)$$

As the cardinality axiom (7) is equivalent to $\binom{m}{k-1}$ disjunctions, a parity axiom of the form of (10) or (11) is in general equivalent to $2^{k-1}$ Boolean disjunctions.

### 4.2 Formalizing Structure

Of course, the ground axiomatizations (6), (9) and (12) are equivalent to the original descriptions given by (5), (7) and (11), so that any structure present in these original descriptions is still there. That structure has, however, been obscured by the ground encodings. Our goal in this section is to begin the process of understanding the structure in a way that lets us describe it in general terms.

As a start, note that each of the axiom sets consists of axioms of equal length; it follows that the axioms can all be obtained from a single one simply by permuting the literals in the theory. In (6) and (9), literals are permuted with other literals of the same sign; in (12), literals are permuted with their negated versions. But in every instance, a permutation suffices.

Thus, for example, the set of permutations needed to generate (9) from the first ground axiom alone is clearly just the set

$$\Omega = \text{Sym}(\{x_1, x_2, x_3, x_4, x_5\}) \tag{13}$$

since these literals can be permuted arbitrarily to move from one element of (9) to another. The set $\Omega$ in (13) is a subgroup of the full permutation group $S_{2n}$ on $2n$ literals in $n$ variables, since $\Omega$ is easily seen to be closed under inversion and composition.

What about the example (12) involving a parity clause? Here the set of permutations needed to generate the four axioms from the first is given by:

$$(x_1, \neg x_1)(x_2, \neg x_2) \tag{14}$$
$$(x_1, \neg x_1)(x_3, \neg x_3) \tag{15}$$
$$(x_2, \neg x_2)(x_3, \neg x_3) \tag{16}$$

Literals are now being exchanged with their negations, but this set, too, is closed under the group inverse and composition operations. Since each element is a composition of disjoint transpositions, each element is its own inverse. The composition of the first two elements is the third.

The remaining example (6) is a bit more subtle; perhaps this is to be expected, since the axiomatization (6) obscures the underlying structure far more effectively than does either (9) or (12).

To understand this example, note that the set of axioms (6) is "generated" by a set of transformations on the underlying variables. In one transformation, we swap the values of $A$ and $B$ for $x$ while leaving the values for $y$ and $z$ unchanged, corresponding to the permutation

$$(a(A,A), a(B,A))(a(A,B), a(B,B))(c(A,A), c(B,A))(c(A,B), c(B,B))$$

We have included in a single permutation the induced changes to all of the relevant ground literals. (The relation $b$ doesn't appear because $b$ does not have $x$ as an argument in (5).) In terms of the literals in (6), this becomes

$$\omega_x = (l_1 l_3)(l_2 l_4)(l_9 l_{11})(l_{10} l_{12})$$

In a similar way, swapping the two values for $y$ corresponds to the permutation

$$\omega_y = (l_1 l_2)(l_3 l_4)(l_5 l_7)(l_6 l_8)$$

and $z$ produces

$$\omega_z = (l_5 l_6)(l_7 l_8)(l_9 l_{10})(l_{11} l_{12})$$

Now consider $\Omega = \langle \omega_x, \omega_y, \omega_z \rangle$, the subgroup of $\text{Sym}(\{l_1, \ldots, l_{12}\})$ that is generated by $\omega_x$, $\omega_y$ and $\omega_z$. Since the clauses in (6) can be obtained from any single clause by permuting the values of $x$, $y$ and $z$, it is clear that the image of any single clause in the set (6) under $\Omega$ is exactly the complete set of clauses (6).

As an example, operating on the first axiom in (6) with $\omega_x$ produces

$$l_3 \vee l_5 \vee l_{11}$$

This is the fifth axiom, as it should be, since we have swapped $a(A, A)$ with $a(B, A)$ and $c(A, A)$ with $c(B, A)$.

Alternatively, a straightforward calculation shows that

$$\omega_x \omega_y = (l_1 l_4)(l_2 l_3)(l_5 l_7)(l_6 l_8)(l_9 l_{11})(l_{10} l_{12})$$

and maps the first axiom in (9) to the next-to-last, the second axiom to last, and so on.

It should be clear at this point what all of these examples have in common. In every case, the set of ground instances corresponding to a single non-Boolean axiom can be generated from any *single* ground instance by the elements of a subgroup of the group $S_{2n}$ of permutations of the literals in the problem.

Provided that all of the clauses are the same length, there is obviously some sub*set* (as opposed to sub*group*) of $S_{2n}$ that can produce all of the clauses from a single one. But subgroups are highly structured objects; there are many fewer subgroups of $S_{2n}$ than there are subsets.[4] One would not expect, a priori, that the particular sets of permutations arising in our examples would all have the structure of subgroups. The fact that they do, that all of these particular subsets are subgroups even though so few subsets are in general, is what leads to our general belief that the structure of the subgroups captures and generalizes the general idea of structure underlying our motivating examples.

In problems without structure, the subgroup property is absent. An instance of random 3-SAT, for example, can always be encoded using a single 3-literal clause $c$ and then that set of permutations needed to recover the entire problem from $c$ in isolation. There is no structure to the set of permutations because the original set of clauses was itself unstructured. In the examples we have been considering, on the other hand, the structure is implicit in the requirement that the set $\Omega$ used to produce the clauses be a group. As we will see, this group structure also has just the computational properties needed if we are to lift RBL and other Boolean satisfiability techniques to our broader setting.

Let us also point out the surprising fact that the subgroup idea captures *all* of the structures discussed in ZAP1. It is not surprising that the various structures used to reduce proof size all have a similar flavor, or that the structure used to speed unit propagation be uniform. But it strikes us as remarkable that these two types of structure, used for such different purposes, are in fact instances of a single framework.

---

4. In general, $S_{2n}$ has $2^{(2n)!}$ subsets, of which only approximately $2^{n^2/4}$ are subgroups (Pyber, 1993).

This, then, is the technical insight on which ZAP rests: Instead of generalizing the language of Boolean satisfiability as seems required by the range of examples we have considered, it suffices to annotate ground clauses with the $\Omega$ needed to reproduce a larger axiom set. Before we formalize this, however, note that any "reasonable" permutation that maps a literal $l_1$ to another literal $l_2$ should respect the semantics of the axiomatization and map $\neg l_1$ to $\neg l_2$ as well.

**Definition 4.1** *Given a set of $n$ variables, we will denote by $W_n$ that subgroup of $S_{2n}$ consisting of permutations that map the literal $\neg l_1$ to $\neg l_2$ whenever they map $l_1$ to $l_2$.*

Informally, an element of $W_n$ corresponds to a permutation of the $n$ variables, together with a choice to flip some subset of them; $W_n$ is therefore of size $|W_n| = 2^n n!$.[5]

We are now in a position to state:

**Definition 4.2** *An* augmented clause *in an $n$-variable Boolean satisfiability problem is a pair $(c, G)$ where $c$ is a Boolean clause and $G \leq W_n$. A ground clause $c'$ is an* instance *of an augmented clause $(c, G)$ if there is some $g \in G$ such that $c' = c^g$. The clause $c$ will be called the* base instance *of $(c, G)$.*

Our aim in the remainder of this paper is to show that augmented clauses have the properties needed to justify the claims we made in the introduction:

1. **They can be represented compactly,**

2. **They can be combined efficiently using a generalization of resolution,**

3. **They generalize existing concepts such as quantification over finite domains, cardinality, and parity clauses, together with providing natural generalizations for proof techniques involving such clauses,**

4. **RBL can be extended with little or no computational overhead to manipulate augmented clauses instead of ground ones, and**

5. **Propagation can be computed efficiently in this generalized setting.**

The first four points will be discussed in this and the next three sections of the paper. The final point is presented in the next paper in this series.

### 4.3 Efficiency of Representation

For the first point, the fact that the augmentations $G$ can be represented compactly is a consequence of $G$'s group structure. In the example surrounding the reconstruction of (9) from (13), for example, the group in question is the full symmetry group on $m$ elements, where $m$ is the number of variables in the cardinality clause. In the lifting example (12),

---

5. We note in passing that $W_n$ is the so-called wreath product of $S_2$ and $S_n$, typically denoted $S_2 \wr S_n$. The specific group $W_n$ is also called the group of "permutations and complementations" by Harrison (1989).

we can describe the group in terms of the generators $\omega_x$, $\omega_y$ and $\omega_z$ instead of listing all eight elements that the group contains. In general, we have (recall that proofs appear in the appendix):

**Proposition 4.3** *Let $S$ be a set of ground clauses, and $(c, G)$ an equivalent augmented clause, where $G$ is represented by generators. It is possible in polynomial time to find a set of generators $\{\omega_1, \ldots, \omega_k\}$ where $k \leq \log_2 |G|$ and $G = \langle \omega_1, \ldots, \omega_k \rangle$.*

Since the size of the full permutation group $S_n$ is only $n! < n^n$ and a single generator takes at most $O(n)$ space, we have:

**Corollary 4.4** *Any augmented clause in a theory containing $n$ literals can be expressed in $O(n^2 \log_2 n)$ space.* □

This result can be strengthened using:

**Proposition 4.5 (Jerrum, 1986; Knuth, 1991)** *Let $G \leq S_n$. It is possible to find in polynomial time a set of generators for $G$ of size at most $O(n)$.* □

This reduces the $O(n^2 \log_2 n)$ in the corollary to simply $O(n^2)$.[6]

Before proceeding, let us make a remark regarding computational complexity. All of the group-theoretic constructs of interest to us can be computed in time polynomial in the group size; basically one simply enumerates the group and evaluates the construction (generate and test, as it were). What is interesting is the collection of group constructions that can be computed in time polynomial in the number of *generators* of the group and the number of variables in the problem. Given Proposition 4.5, the time is thus polynomial in the number of variables in the problem.

Note that the size of the group $G$ can be vastly greater than the number of instances of any particular augmented clause $(c, G)$. As an example, for the cardinality clause

$$x_1 + \cdots + x_m \geq k \tag{17}$$

the associated symmetry group $\text{Sym}\{x_1, \ldots, x_m\}$ acts on an instance such as

$$x_1 \vee \cdots \vee x_{m-k+1} \tag{18}$$

to reproduce the full Boolean axiomatization. But each such instance corresponds to $(m - k + 1)!$ distinct group elements as the variables within the clause (18) are permuted.

In this particular case, the symmetry group $\text{Sym}\{x_1, \ldots, x_m\}$ can in fact be generated by the two permutations $(x_1, x_2)$ and $(x_2, x_3, \ldots, x_m)$.

**Definition 4.6** *Two augmented clauses $(c_1, G_1)$ and $(c_2, G_2)$ will be called* equivalent *if they have identical sets of instances. This will be denoted $(c_1, G_1) \equiv (c_2, G_2)$.*

---

6. Although the methods used are nonconstructive, Babai (1986) showed that the length of an increasing sequence of subgroups of $S_n$ is at most $\lfloor \frac{3n}{2} \rfloor - 2$; this imposes the same bound on the number of generators needed (compare the proof of Proposition 4.3). Using other methods, McGiver and Neumann stated (1987) that for $n \neq 3$, there is always a generating set of size at most $\lfloor \frac{n}{2} \rfloor$.

**Proposition 4.7** *Let $(c, G)$ be an augmented clause. Then if $c'$ is any instance of $(c, G)$, $(c, G) \equiv (c', G)$.*

We also have:

**Proposition 4.8** *Let $(c, G)$ be an augmented clause with $d$ distinct instances. Then there is a subgroup $H \leq G$ that can be described using $O(\log_2(d))$ generators such that $(c, H) \equiv (c, G)$. Furthermore, given $d$ and generators for $G$, there is a Monte Carlo polynomial-time algorithm for constructing the generators of such an $H$.*[7]

Proposition 4.5 is the first of the results promised in the introduction: If $d$ Boolean axioms involving $n$ variables can be captured as instances of an augmented clause, that augmented clause can be represented using $O(n)$ generators; Proposition 4.8 guarantees that $O(\log_2 d)$ generators suffice as well.

In the specific instances that we have discussed, the representational efficiencies are greater still:

| clause type | Boolean axioms | generators | total size |
|---|---|---|---|
| cardinality | $\binom{m}{k-1}$ | 2 | $m+1$ |
| parity | $2^{k-1}$ | 3 | $k+5$ |
| QPROP | $d^v$ | $2v$ | $v(d+1)$ |

Each row gives the number of Boolean axioms or generators needed to represent a clause of the given type, along with the total size of those generators. For the cardinality clause (17), the complete symmetry group over $m$ variables can be expressed using exactly two generators, one of size 2 and the other of size $m-1$.[8] The number of Boolean axioms is $\binom{m}{k-1}$ as explained in Section 4.1.2.

For the parity clause
$$x_1 + \cdots + x_k \equiv m \pmod{2}$$
the number of Boolean axioms is the same as the number of ways to select an even number of the $x_i$'s, which is half of all of the subsets of $\{x_1, \ldots, x_k\}$. (Remove $x_1$ from the set; now *any* subset of the remaining $x_i$ can be made of even parity by including $x_1$ or not as appropriate.) The parity groups $F_k$ can be captured by $k-1$ generators of the form $(x_1, \neg x_1), (x_i, \neg x_i)$ as $i = 2, \ldots, k$ (total size $4(k-1)$); alternatively, one can combine the single generator $(x_1, \neg x_1)(x_2, \neg x_2)$ with the full symmetry group on $x_1, \ldots, x_k$ to describe a parity clause using exactly three generators (total size $4 + 2 + k - 1$).

Finally, a QPROP clause involving $v$ variables, each with a domain of size $d$, corresponds to a set of $d^v$ individual domain axioms. As we saw in Section 4.2 and will formalize in Section 6.1, the associated group can be described using symmetry groups over the domains of each quantified variable; there are $v$ such groups and two generators (of size 2 and $d-1$) are required for each.[9]

---

7. A Monte Carlo algorithm is one that is not deterministic but that can be made to work with arbitrarily high specified probability without changing its overall complexity (Seress, 2003).
8. As noted earlier, $S_n$ is generated by the transposition $(1, 2)$ and the $n-1$-cycle $(2, 3, \ldots, n)$.
9. Depending on the sizes, the number of generators needed for a product of symmetry groups can be reduced in many cases, although the total size is unchanged.

Note that the total sizes are virtually optimal in all of these cases. For cardinality and parity clauses, it is surely essential to enumerate the variables in question (size $m$ and $k$ respectively). For QPROP clauses, simply enumerating the domains of quantification takes space $vd$.

## 5. Resolution

We now turn to the question of basing derivations on augmented clauses instead of ground ones. We begin with a few preliminaries:

**Proposition 5.1** *For ground clauses $c_1$ and $c_2$ and a permutation $\omega \in W_n$,*

$$\texttt{resolve}(\omega(c_1), \omega(c_2)) = \omega(\texttt{resolve}(c_1, c_2))$$

**Definition 5.2** *If $C$ is a set of augmented clauses, we will say that $C$ entails an augmented clause $(c, G)$, writing $C \models (c, G)$, if every instance of $(c, G)$ is entailed by the set of instances of the augmented clauses in $C$.*

We are now in a position to consider lifting the idea of resolution to our setting, but let us first discuss the overall intent of this lifting. What we would like to do is to think of an augmented clause as having force similar to all of its instances; as a result, when we resolve two augmented clauses $(c_1, G_1)$ and $(c_2, G_2)$, we would like to obtain as the (augmented) resolvent the set of all resolutions that are sanctioned by resolving an instance of $(c_1, G_1)$ with one of $(c_2, G_2)$. Unfortunately, we have:

**Proposition 5.3** *There are augmented clauses $c_1$ and $c_2$ such that the set $S$ of resolvents of instances of the two clauses does not correspond to any single augmented clause $(c, G)$.*

Given that we cannot capture *exactly* the set of possible resolvents of two augmented clauses, what *can* we do? If $(c, G)$ is an "augmented resolvent" of $(c_1, G_1)$ and $(c_2, G_2)$, we might expect $(c, G)$ to have the following properties:

1. It should be *sound*, in that $(c_1, G_1) \wedge (c_2, G_2) \models (c, G)$. This says that every instance of $(c, G)$ is indeed sanctioned by resolving an instance of $(c_1, G_1)$ with an instance of $(c_2, G_2)$.

2. It should be *complete*, in that $\texttt{resolve}(c_1, c_2)$ is an instance of $(c, G)$. We may not be able to capture every possible resolvent in the augmented clause $(c, G)$, but we should surely capture the base case that is obtained by resolving the base instance $c_1$ directly against the base instance $c_2$.

3. It should be *monotonic*, in that if $G_1 \leq H_1$ and $G_2 \leq H_2$, then $(c, G)$ is also a resolvent of $(c_1, H_1)$ and $(c_2, H_2)$. As the clauses being resolved become stronger, the resolvent should become stronger as well.

4. It should be *polynomial*, in that it is possible to confirm that $(c, G)$ is a resolvent of $(c_1, G_1)$ and $(c_2, G_2)$ in polynomial time.

5. It should be *stable*, in that if $c_1^G = c_2^G$, then $(\texttt{resolve}(c_1, c_2), G)$ is a resolvent of $(c_1, G)$ and $(c_2, G)$. Roughly speaking, this says that if the groups in the two input clauses are the same, then the augmented resolvent can be obtained by resolving the base instances $c_1$ and $c_2$ and then operating with the same group.

6. It should be *strong*, in that if no element of $c_1^{G_1}$ is moved by $G_2$ and similarly no element of $c_2^{G_2}$ is moved by $G_1$, then $(\texttt{resolve}(c_1, c_2), \langle G_1, G_2 \rangle)$ is a resolvent of $(c_1, G_1)$ and $(c_2, G_2)$. This says that if the group actions are distinct in that $G_1$ acts on $c_1$ and leaves $c_2$ completely alone and vice versa, then we should be able to get the complete group of resolvents in our answer. This group corresponds to be base resolvent $\texttt{resolve}(c_1, c_2)$ acted on by the group generated by $G_1$ and $G_2$.

**Definition 5.4** *A definition of augmented resolution will be called* satisfactory *if it satisfies the above conditions.*

Note that we do *not* require that the definition of augmented resolution be unique. Our goal is to define conditions under which $(c, G)$ is *an* augmented resolvent of $(c_1, G_1)$ and $(c_2, G_2)$, as opposed to "the" augmented resolvent of $(c_1, G_1)$ and $(c_2, G_2)$. To the best of our knowledge (and as suggested by Proposition 5.3), there is no satisfactory definition of augmented resolution that defines the resolvent of two augmented clauses uniquely.

As we work toward a satisfactory definition of augmented resolution, let us consider some examples to help understand what the basic issues are. Consider, for example, resolving the two clauses

$$(a \vee b, \langle (bc) \rangle)$$

which has instances $a \vee b$ and $a \vee c$ and

$$(\neg a \vee d, \langle \rangle)$$

which has the single instance $\neg a \vee d$. We will write these somewhat more compactly as

$$(a \vee b, (bc)) \tag{19}$$

and

$$(\neg a \vee d, \mathbf{1}) \tag{20}$$

respectively.

Resolving the clauses individually, we see that we should be able to derive the pair of clauses $b \vee d$ and $c \vee d$; in other words, the augmented clause

$$(b \vee d, (bc)) \tag{21}$$

It certainly seems as if it should be possible to capture this in our setting, since the base instance of (21) is just the resolvent of the base instances of (19) and (20).[10] Where does the group generated by $(bc)$ come from?

To answer this, we begin by introducing some additional notation.

---

10. Indeed, condition (6) requires that $b \vee d$ be an instance of some augmented resolvent, since the groups act independently in this case.

**Definition 5.5** *Let $\omega$ be a permutation and $S$ a set. Then by $\omega|_S$ we will denote the result of restricting the permutation to the given set.*

Note that $\omega|_S$ will be a permutation on $S$ if and only if $S^\omega = S$, so that $S$ is fixed by $\omega$.

**Definition 5.6** *For $K_1, \ldots, K_n \subseteq L$ and $G_1, \ldots, G_n \leq \mathrm{Sym}(L)$, we will say that a permutation $\omega \in \mathrm{Sym}(L)$ is an extension of $\{G_1, \ldots, G_n\}$ for $\{K_1, \ldots, K_n\}$ if there are $g_i \in G_i$ such that for all $i$, $\omega|_{K_i} = g_i|_{K_i}$. We will denote the set of such extensions by $\mathtt{extn}(K_i, G_i)$.*

The definition says that any particular extension $x \in \mathtt{extn}(K_i, G_i)$ must simultaneously extend elements of all of the individual groups $G_i$, when those groups act on the various subsets $K_i$.

As an example, suppose that $K_1 = \{a, b\}$ and $K_2 = \{\neg a, e\}$, with $G_1 = \mathrm{Sym}\{b, c, d\}$ and $G_2 = \langle (ed) \rangle$. A permutation is an extension of the $G_i$ for $K_i$ if and only if, when restricted to $\{a, b\}$ it is a member of $\mathrm{Sym}(b, c, d)$ and, when restricted to $\{a, e\}$ it is a member of $\langle (ed) \rangle$. In other words, $b$ can be mapped to $b$, $c$ or $d$, and $e$ can be mapped to $d$ if desired. The set of extensions is thus

$$\{(), (bc), (bd), (bcd), (bdc), (cd), (ed), (edc), (bc)(de)\} \tag{22}$$

Note that this set is not a group because it is not closed under the group operations; $(edc)$ is permitted ($e$ is mapped to $d$ and we don't care where $d$ and $c$ go), but $(edc)^2 = (ecd)$ is not.

Considered in the context of resolution, suppose that we are trying to resolve augmented clauses $(c_1, G_1)$ and $(c_2, G_2)$. At some level, any result $(\mathtt{resolve}(c_1, c_2), G)$ for which $G \subseteq \mathtt{extn}(c_i, G_i)$ should be a legal resolvent of the original clauses, since each instance is sanctioned by resolving instances of the originals.[11] We can't define the resolvent to be $(\mathtt{resolve}(c_1, c_2), \mathtt{extn}(c_i, G_i))$ because, as we have seen from our example, neither $\mathtt{extn}(c_i, G_i)$ nor $\mathtt{extn}(c_i, G_i) \cap W_n$ need be a group. (We also know this from Proposition 5.3; there may be no single group that captures all of the possible resolvents.) But we can't simply require that the augmented resolvent $(c, G)$ have $G \subseteq \mathtt{extn}(c_i, G_i)$, because there is no obvious polynomial test for inclusion of a group in a set.[12]

To overcome these difficulties, we need a version of Definition 5.6 that produces a *group* of extensions, as opposed to just a set:

**Definition 5.7** *For $K_1, \ldots, K_n \subseteq L$ and $G_1, \ldots, G_n \leq \mathrm{Sym}(L)$, we will say that a permutation $\omega \in \mathrm{Sym}(L)$ is a stable extension of $\{G_1, \ldots, G_n\}$ for $\{K_1, \ldots, K_n\}$ if there are $g_i \in G_i$ such that for all $i$, $\omega|_{K_i^{G_i}} = g_i|_{K_i^{G_i}}$. We will denote the set of stable extensions of $\{G_1, \ldots, G_n\}$ for $\{K_1, \ldots, K_n\}$ by $\mathtt{stab}(K_i, G_i)$.*

---

11. We can't write $G \leq \mathtt{extn}(c_i, G_i)$ because $\mathtt{extn}(c_i, G_i)$ may not be a group.
12. It is possible to test in polynomial time if $G \leq H$, since we can simply test each generator of $G$ for membership in $H$. But if $H$ is not closed under the group operation, the fact that the generators are all in $H$ is not sufficient to conclude that $G \subseteq H$.

This definition is modified from Definition 5.6 only in that the restriction of $\omega$ is not just to the original sets $K_i$ for each $i$, but to $K_i^{G_i}$, the $G_i$-closure of $K_i$ (recall Definition 3.8).

In our example where $K_1 = \{a, b\}$ and $K_2 = \{\neg a, e\}$, with $G_1 = \text{Sym}\{b, c, d\}$ and $G_2 = \langle (ed) \rangle$, a stable extension must be a member of $\text{Sym}(b, c, d)$ when restricted to $\{a, b, c, d\}$ (the $G_1$-closure of $K_1$), and must be a member of $\langle (ed) \rangle$ when restricted to $\{d, e\}$. This means that we *do* care where a candidate permutation maps $c$ and $d$, so that the set of stable extensions, instead of (22), is instead simply $\{(), (bc)\} = \langle (bc) \rangle$. The fact that $d$ has to be mapped to $b$, $c$, or $d$ by virtue of $G_1$ and has to be mapped to either $d$ or $e$ by virtue of $G_2$ means that $d$ has to be fixed by any permutation in $\texttt{stab}(K_i, G_i)$, which is why the resulting set of stable extensions is so small.

In general, we have:

**Proposition 5.8** $\texttt{stab}(K_i, G_i) \leq \text{Sym}(L)$.

In other words, $\texttt{stab}(K_i, G_i)$ is a subgroup of $\text{Sym}(L)$.

At this point, we still need to deal with the monotonicity condition (3) of Definition 5.4. After all, if we have
$$(c, G) \models (c', G')$$
we should also have
$$\texttt{resolve}((c, G), (d, H)) \models \texttt{resolve}((c', G'), (d, H))$$

To see why this is an issue, suppose that we are resolving
$$(a \vee b, \text{Sym}\{b, c, d\}) \tag{23}$$
with
$$(\neg a \vee e, (ed)) \tag{24}$$
Because the groups both act on $d$, we have already seen that if we take the group of stable extensions as the group in the resolvent, we will conclude $(b \vee e, (bc))$. But if we replace (24) with $(\neg a \vee e, \mathbf{1})$ before resolving, the result is the stronger $(b \vee e, \text{Sym}\{b, c, d\})$. If we replace (23) with $(a \vee b, (bc))$, the result is the different but also stronger $(b \vee e, \langle (bc), (de) \rangle)$

The monotonicity considerations suggest:

**Definition 5.9** *Suppose that $(c_1, G_1)$ and $(c_2, G_2)$ are augmented clauses where $c_1$ and $c_2$ resolve in the conventional sense. Then a resolvent of $(c_1, G_1)$ and $(c_2, G_2)$ is any augmented clause of the form $(\texttt{resolve}(c_1, c_2), G)$ where $G \leq \texttt{stab}(c_i, H_i) \cap W_n$ for some $H_i \leq G_i$ for $i = 1, 2$. The canonical resolvent of $(c_1, G_1)$ and $(c_2, G_2)$, to be denoted by $\texttt{resolve}((c_1, G_1), (c_2, G_2))$, is the augmented clause $(\texttt{resolve}(c_1, c_2), \texttt{stab}(c_i, G_i) \cap W_n)$.*

**Proposition 5.10** *Definition 5.9 of (noncanonical) resolution is satisfactory. The definition of canonical resolution satisfies all of the conditions of Definition 5.4 except for monotonicity.*

Before proceeding, let us consider the example that preceded Definition 5.9 in a bit more detail. There, we were looking for augmented resolvents of $(a \vee b, \mathrm{Sym}\{b,c,d\})$ from (23) and $(\neg a \vee e, (ed))$ from (24).

To find such a resolvent, we begin by selecting $H_1 \leq G_1 = \mathrm{Sym}\{b,c,d\}$ and $H_2 \leq G_2 = \langle (ed) \rangle$. We then need to use Definition 5.7 to compute the group of stable extensions of $(c_1, H_1)$ and $(c_2, H_2)$.

If we take $H_1$ and $H_2$ to be the trivial group $\mathbf{1}$, then the group of stable extensions is also trivial, so we see that

$$(\mathtt{resolve}(a \vee b, \neg a \vee e), \mathbf{1}) = (b \vee e, \mathbf{1})$$

is a resolvent of (23) and (24). Other choices for $H_1$ and $H_2$ are more interesting.

If we take $H_1 = \mathbf{1}$ and $H_2 = G_2$, the stable extensions leave the first clause fixed but can move the image of the second consistent with $G_2$. This produces the augmented resolvent $(b \vee e, (de))$.

If, on the other hand, we take $H_1 = G_1$ and $H_2 = \mathbf{1}$, we have to leave $e$ fixed but can exchange $b$, $c$ and $d$ freely, and we get $(b \vee e, \mathrm{Sym}\{b,c,d\})$ as the resolvent.

If $H_1 = G_1$ and $H_2 = G_2$, we have already computed the group of stable extensions in earlier discussions of this example; the augmented resolvent is $(b \vee e, (bc))$, which is weaker than the resolvent of the previous paragraph. And finally, if we take $H_1 = \langle (bc) \rangle$ and $H_2 = G_2$, we can exchange $b$ and $c$ or independently exchange $d$ and $e$ so that we get the augmented resolvent $(b \vee e, \langle (bc), (de) \rangle)$. These choices have already been mentioned in the discussion of monotonicity that preceded Definition 5.9.

There is a variety of additional remarks to be made about Definition 5.9. First, *canonical* resolution lacks the monotonicity property, as shown by our earlier example. In addition, the resolvent of two augmented clauses can obviously depend on the choice of the representative elements in addition to the choice of subgroup of $\mathtt{stab}(c_i, G_i)$. Thus, if we resolve

$$(l_1, (l_1 l_2)) \tag{25}$$

with

$$(\neg l_1, \mathbf{1}) \tag{26}$$

we get a contradiction. But if we rewrite (25) so that we are attempting to resolve (26) with

$$(l_2, (l_1 l_2))$$

no resolution is possible at all.

To address this in a version of RBL that has been lifted to our more general setting, we need to ensure that if we are trying to resolve $(c_1, G_1)$ and $(c_2, G_2)$, the base instances $c_1$ and $c_2$ themselves resolve. As we will see, this can be achieved by maintaining ground reasons for each literal in the annotated partial assignment. These ground clauses will always resolve when a contradiction is found and the search needs to backtrack.

We should also point out that there are computational issues involved in the evaluation of $\mathtt{stab}(c_i, G_i)$. If the component groups $G_1$ and $G_2$ are described by listing their elements, an incremental construction is possible where generators are gradually added until it is impossible to extend the group further without violating Definition 5.9. But if $G_1$ and $G_2$ are described only in terms of their generators, as suggested by the results in Section 4.3, computing $\text{stab}(c_i, G_i)$ involves the following computational subtasks (Dixon et al., 2004a) (recall Definition 3.9):[13]

1. Given a group $G$ and set $C$, find $G_{\{C\}}$.

2. Given a group $G$ and set $C$, find $G_C$.

3. Given two groups $G_1$ and $G_2$ described in terms of generators, find a set of generators for $G_1 \cap G_2$.

4. Given $G$ and $C$, let $\omega \in G_{\{C\}}$. Now $\omega|_C$, the restriction of $\omega$ to $C$, makes sense because $C^\omega = C$. Given a $\rho$ that is such a restriction, find an element $\rho' \in G$ such that $\rho'|_C = \rho$.

We will have a great deal more to say about these issues in the paper describing the ZAP implementation. At this point, we remark merely that time complexity is known to be polynomial only for the second and fourth of the above problems; the other two are not known to be in polynomial time. However, computational group theory systems incorporate procedures that rarely exhibit super-polynomial behavior even though one can construct examples that force them to take exponential time (as usual, in terms of the number of generators of the groups, not their absolute size).

In the introduction, we claimed that the result of resolution was unique using reasons and that ZAP's fundamental inference step was both in NP with respect to the ZAP representation and of low-order polynomial complexity in practice. The use of reasons breaks the ambiguity surrounding (25) and (26), and the remarks regarding complexity correspond to the computational observations just made.

## 6. Examples and Proof Complexity

Let us now turn to the examples that we have discussed previously: first-order axioms that are quantified over finite domains, along with the standard examples from proof complexity, including pigeonhole problems, clique coloring problems and parity clauses. For the first, we will see that our ideas generalize conventional notions of quantification while providing additional representational flexibility in some cases. For the other examples, we will present a ground axiomatization, recast it using augmented clauses, and then give a polynomially sized derivation of unsatisfiability using augmented resolution.

### 6.1 Lifted Clauses and QPROP

To deal with lifted clauses, suppose that we have a quantified clause such as

$$\forall xyz.a(x,y) \vee b(y,z) \vee c(z) \tag{27}$$

---

13. There is an additional requirement that we be able to compute $\text{stab}(c_i, G_i) \cap W_n$ from $\text{stab}(c_i, G_i)$. This is not an issue in practice because we work with an overall representation in which all groups are represented by their intersections with $W_n$. Thus if $g$ is included as a generator for a group $G$, we automatically include in the generators for $G$ the "dual" permutation to $g$ where every literal has had its sign flipped.

Suppose that the domain of $x$ is $X$, that of $y$ is $Y$, and that of $z$ is $Z$. Thus a grounding of the clause (27) involves working with a map that takes three elements $x \in X$, $y \in Y$ and $z \in Z$ and produces the ground atoms corresponding to $a(x,y)$, $b(y,z)$ and $c(z)$. In other words, if $V$ is the set of variables in our problem, there are injections

$$\begin{aligned} a &: X \times Y \to V \\ b &: Y \times Z \to V \\ c &: Z \to V \end{aligned}$$

where the images of $a$, $b$ and $c$ are disjoint and each is an injection because distinct relation instances must be mapped to distinct ground atoms.

Now given a permutation $\omega$ of the elements of $X$, we can define a permutation $\rho_X(\omega)$ on $V$ given by:

$$\rho_X(\omega)(v) = \begin{cases} a(\omega(x), y), & \text{if } v = a(x,y); \\ v; & \text{otherwise.} \end{cases}$$

In other words, there is a mapping $\rho_X$ from the set of permutations on $X$ to the set of permutations on $V$:

$$\rho_X : \text{Sym}(X) \to \text{Sym}(V)$$

**Definition 6.1** *Let $G$ and $H$ be groups and $f : G \to H$ a function between them. $f$ will be called a* homomorphism *if it respects the group operation in that $f(g_1 g_2) = f(g_1) f(g_2)$.*

It should be clear that:

**Proposition 6.2** $\rho_X : \text{Sym}(X) \to \text{Sym}(V)$ *is an injection and a homomorphism.* □

In other words, $\rho_X$ makes a "copy" of $\text{Sym}(X)$ inside of $\text{Sym}(V)$ corresponding to permuting the elements of $x$'s domain $X$.

In a similar way, we can define homomorphisms $\rho_Y$ and $\rho_Z$ given by

$$\rho_Y(\omega)(v) = \begin{cases} a(x, \omega(y)), & \text{if } v = a(x,y); \\ b(\omega(y), z), & \text{if } v = b(y,z); \\ v; & \text{otherwise.} \end{cases}$$

and

$$\rho_Z(\omega)(v) = \begin{cases} b(y, \omega(z)), & \text{if } v = b(y,z); \\ c(\omega(z)), & \text{if } v = c(z); \\ v; & \text{otherwise.} \end{cases}$$

Now consider the subgroup of $\text{Sym}(V)$ generated by the three images $\rho_X(\text{Sym}(X))$, $\rho_Y(\text{Sym}(Y))$ and $\rho_Z(\text{Sym}(Z))$. It is clear that the three images commute with one another, and that their intersection is only the trivial permutation. This means that $\rho_X$, $\rho_Y$ and $\rho_Z$ collectively inject the product $\text{Sym}(X) \times \text{Sym}(Y) \times \text{Sym}(Z)$ into $\text{Sym}(V)$; we will denote this by

$$\rho_{XYZ} : \text{Sym}(X) \times \text{Sym}(Y) \times \text{Sym}(Z) \to \text{Sym}(V)$$

and it should be clear that the original quantified axiom (27) is equivalent to the augmented axiom

$$(a(A,B) \vee b(B,C) \vee c(C), \rho_{XYZ}(\text{Sym}(X) \times \text{Sym}(Y) \times \text{Sym}(Z)))$$

where $A$, $B$ and $C$ are any (not necessarily distinct) elements of $X$, $Y$ and $Z$, respectively. The quantification is exactly captured by the augmentation.

The interesting thing is what happens to resolution in this setting:

**Proposition 6.3** *Let $p$ and $q$ be quantified clauses such that there is a term $t_p$ in $p$ and $\neg t_q$ in $q$ where $t_p$ and $t_q$ have common instances. Suppose also that $(p_g, P)$ is an augmented clause equivalent to $p$ and $(q_g, Q)$ is an augmented clause equivalent to $q$, where $p_g$ and $q_g$ resolve. Then if no terms in $p$ and $q$ except for $t_p$ and $t_q$ have common instances, the result of resolving $p$ and $q$ in the conventional lifted sense is equivalent to* $\texttt{resolve}((p_g, P), (q_g, Q))$.

Here is an example. Suppose that $p$ is

$$a(A, x) \lor b(C, y, z) \lor c(x, y, z) \tag{28}$$

and $q$ is

$$a(B, x) \lor \neg b(x, D, z) \tag{29}$$

so that the most general unifier of the two appearances of $b$ binds $x$ to $C$ in (29) and $y$ to $D$ in (28) to produce

$$a(A, x) \lor c(x, D, z) \lor a(B, C) \tag{30}$$

In the version using augmented clauses, it is clear that if we select ground instances $p_g$ of (29) and $q_g$ of (28) that resolve, the resolvent will be a ground instance of (30); the interesting part is the group. For this, note simply that the image of $p_g$ is the entire embedding of $\text{Sym}(X) \times \text{Sym}(Y) \times \text{Sym}(Z)$ into $\text{Sym}(L)$ corresponding to the lifted axiom (28), and the image of $q_g$ is similarly the embedded image of $\text{Sym}(X) \times \text{Sym}(Z)$ corresponding to (29).

The group of stable extensions of the two embeddings corresponds to any bindings for the variables in (28) and (29) that can be extended to a permutation of all of the variables in question; in other words, to bindings that (a) are consistent in that common ground literals in the two expressions are mapped to the same ground literal by both sets of bindings, and (b) are disjoint in that we do not attempt to map two quantified literals to the same ground instance. This latter condition is guaranteed by the conditions of the proposition, which require that the non-resolving literals have no common ground instances. In this particular example, if we choose the instances

$$a(A, \alpha) \lor b(C, D, \beta) \lor c(\alpha, D, \beta)$$

for (28) and

$$a(B, C) \lor \neg b(C, D, \beta)$$

for (29), the resulting augmented clause is

$$(a(A, \alpha) \lor c(\alpha, D, \beta) \lor a(B, C), G) \tag{31}$$

where $G$ is the group mapping $\text{Sym}(X) \times \text{Sym}(Z)$ into $\text{Sym}(L)$ so that (31) corresponds to the quantified clause (30).

The condition requiring lack of commonality of ground instances is necessary; consider resolving

$$a(x) \lor b$$

with
$$a(y) \vee \neg b$$

In the quantified case, we get
$$\forall xy. a(x) \vee a(y) \tag{32}$$

In the augmented case, it is not hard to see that if we resolve $(a(A) \vee b, G)$ with $(a(A) \vee \neg b, G)$ we get
$$(a(A), G)$$
corresponding to
$$\forall x. a(x) \tag{33}$$

while if we choose to resolve $(a(A) \vee b, G)$ with $(a(B) \vee \neg b, G)$, we get instead
$$\forall x \neq y. a(x) \vee a(y)$$

It is not clear which of these representations is superior. The conventional (32) is more compact, but obscures the fact that the stronger (33) is entailed as well. This particular example is simple, but other examples involving longer clauses and some residual unbound variables can be more complex.

### 6.2 Proof Complexity

We conclude this section with a demonstration that ZAP can produce polynomial proofs of the problem instances appearing in the original table of the introduction.

#### 6.2.1 Pigeonhole Problems

Of the examples known to be exponentially difficult for conventional resolution-based systems, pigeonhole problems are in many ways the simplest. As usual, we will denote by $p_{ij}$ the fact that pigeon $i$ (of $n+1$) is in hole $j$ of $n$, so that there are $n^2 + n$ variables in the problem. We denote by $G$ the subgroup of $W_{n^2+n}$ that allows arbitrary exchanges of the $n+1$ pigeons or the $n$ holes, so that $G$ is isomorphic to $S_{n+1} \times S_n$. This particular example is straightforward because there is a single global group that we will be able to use throughout the analysis.

Our axiomatization is now:
$$(\neg p_{11} \vee \neg p_{21}, G) \tag{34}$$

saying that no two pigeons can be in the same hole, and

$$(p_{11} \vee \cdots \vee p_{1n}, G) \tag{35}$$

saying that the first (and thus every) pigeon has to be in some hole.

**Proposition 6.4** *There is an augmented resolution proof of polynomial size of the mutual unsatisfiability of (34) and (35).*

**Proof.** This is a consequence of the stronger Proposition 6.5 below, but we also present an independent proof in the appendix. The ideas in the proof are needed in the analysis of the clique-coloring problem. ▫

**Proposition 6.5** *Any implementation of Procedure 2.8 that branches on positive literals in unsatisfied clauses on line 12 will produce a proof of polynomial size of the mutual unsatisfiability of (34) and (35), independent of specific branching choices made.*

This strikes us as a remarkable result: Not only is it possible to find a proof of polynomial length in the augmented framework, but in the presence of unit propagation, it is difficult not to!

A careful proof of this result is in the appendix, but it will be useful to examine in detail how a prover might proceed in a small (four pigeon, three hole) example.

We begin by branching on (say) $p_{11}$, saying that pigeon one is in hole one. Now unit propagation allows us to conclude immediately that no other pigeon is in hole one, so our annotated partial assignment is:

| literal | reason |
|---|---|
| $p_{11}$ | `true` |
| $\neg p_{21}$ | $\neg p_{11} \vee \neg p_{21}$ |
| $\neg p_{31}$ | $\neg p_{11} \vee \neg p_{31}$ |
| $\neg p_{41}$ | $\neg p_{11} \vee \neg p_{41}$ |

Now we try putting pigeon two in hole two,[14] and extend the above partial assignment with:

| literal | reason |
|---|---|
| $p_{22}$ | `true` |
| $\neg p_{12}$ | $\neg p_{22} \vee \neg p_{12}$ |
| $\neg p_{32}$ | $\neg p_{22} \vee \neg p_{32}$ |
| $\neg p_{42}$ | $\neg p_{22} \vee \neg p_{42}$ |

At this point, however, we are forced to put pigeons three and four in hole three, which leads to a contradiction $\bot$:

| literal | reason |
|---|---|
| $p_{33}$ | $p_{31} \vee p_{32} \vee p_{33}$ |
| $p_{43}$ | $p_{41} \vee p_{42} \vee p_{43}$ |
| $\bot$ | $\neg p_{33} \vee \neg p_{43}$ |

Resolving the last two reasons produces $\neg p_{33} \vee p_{41} \vee p_{42}$, which we can resolve with the reason for $p_{33}$ to get $p_{41} \vee p_{42} \vee p_{31} \vee p_{32}$. Continuing to backtrack produces $p_{41} \vee \neg p_{22} \vee p_{31}$.

Operating on the clause $p_{41} \vee \neg p_{22} \vee p_{31}$ with the usual symmetry group (swapping hole 2 and hole 3) produces $p_{41} \vee \neg p_{23} \vee p_{31}$, and now there is nowhere for pigeon two to go. We resolve these two clauses with $p_{21} \vee p_{22} \vee p_{23}$ to get $p_{41} \vee p_{31} \vee p_{21}$, and thus $\neg p_{11}$. This implies $\neg p_{ij}$ for all $i$ and $j$ under the usual symmetry, and we conclude that the original axiomatization was unsatisfiable.

---

14. We cannot conclude that pigeon two is in hole two "by symmetry" from the existing choice that pigeon one is in hole one, of course. The symmetry group can only be applied to the original clauses and to derived nogoods, not to branch choices. Alternatively, the branch choice corresponds to the augmented clause $(p_{11}, \mathbf{1})$ and not $(p_{11}, G)$.

6.2.2 CLIQUE COLORING PROBLEMS

The pigeonhole problem is difficult for resolution but easy for many other proof systems; clique coloring problems are difficult for both resolution and other approaches such as pseudo-Boolean axiomatizations (Pudlak, 1997).

Clique coloring problems are derivatives of pigeonhole problems where the exact nature of the pigeonhole problem is obscured. Somewhat more specifically, they say that a graph includes a clique of $n+1$ nodes (where every node in the clique is connected to every other), and that the graph must be colored in $n$ colors. If the graph itself is known to be a clique, the problem is equivalent to the pigeonhole problem. But if we know only that the clique can be embedded into the graph, the problem is more difficult.

To formalize this, we will use $e_{ij}$ to describe the graph, $c_{ij}$ to describe the coloring of the graph, and $q_{ij}$ to describe the embedding of the clique into the graph. The graph has $m$ nodes, the clique is of size $n+1$, and there are $n$ colors available. The axiomatization is:

$$\neg e_{ij} \vee \neg c_{il} \vee \neg c_{jl} \quad \text{for } 1 \leq i < j \leq m, \, l = 1, \ldots, n \tag{36}$$

$$c_{i1} \vee \cdots \vee c_{in} \quad \text{for } i = 1, \ldots, m \tag{37}$$

$$q_{i1} \vee \cdots \vee q_{im} \quad \text{for } i = 1, \ldots, n+1 \tag{38}$$

$$\neg q_{ij} \vee \neg q_{kj} \quad \text{for } 1 \leq i < k \leq n+1, \, j = 1, \ldots, m \tag{39}$$

$$e_{ij} \vee \neg q_{ki} \vee \neg q_{lj} \quad \text{for } 1 \leq i < j \leq m, \, 1 \leq k \neq l \leq n+1 \tag{40}$$

Here $e_{ij}$ means that there is an edge between graph nodes $i$ and $j$, $c_{ij}$ means that graph node $i$ is colored with the $j$th color, and $q_{ij}$ means that the $i$th element of the clique is mapped to graph node $j$. Thus the first axiom (36) says that two of the $m$ nodes in the graph cannot be the same color (of the $n$ colors available) if they are connected by an edge. (37) says that every graph node has a color. (38) says that every element of the clique appears in the graph, and (39) says that no two elements of the clique map to the same node in the graph. Finally, (40) says that the clique is indeed a clique – no two clique elements can map to disconnected nodes in the graph. As in the pigeonhole problems, there is a global symmetry in this problem in that any two nodes, clique elements or colors can be swapped.

**Proposition 6.6** *There is an augmented resolution proof of polynomial size of the unsatisfiability of (36)–(40).*

The proof in the appendix presents a ZAP proof of size $O(m^2 n^2)$ for clique-coloring problems, where $m$ is the size of the graph and $n$ is the size of the clique. The ZAP implementation produces shorter proofs, of size $O((m+n)^2)$ (Dixon et al., 2004a). While short, these proofs involve the derivation and manipulation of subtle clauses and are too complex for us to understand.[15]

Before we move on to parity clauses, note that the approach we are proposing is properly stronger than one based on "symmetry-breaking" axioms (Crawford, Ginsberg, Luks, & Roy, 1996) or the approaches taken by Krishnamurthy (1985) or Szeider (2003). In the symmetry-breaking approach, the original axiom set is modified so that as soon as a single symmetric instance is falsified, so are all of that instance's symmetric variants. Both we and the other authors (Krishnamurthy, 1985; Szeider, 2003) achieve a similar effect by attaching a symmetry to the conclusion; either way, all symmetric instances are removed as soon as it is possible to disprove any. Unlike all of these other authors, however, an approach based on augmented clauses is capable of exploiting local symmetries present in a subset of the entire axiom set. The other authors require the presence of a global symmetry across the entire structure of the problem.

---

15. It is not clear whether one should conclude from this something good about ZAP, or something bad about the authors. Perhaps both.

### 6.2.3 Parity Clauses

Rather than discuss a specific example here, we show that determining the satisfiability of any set of parity clauses is in $P$ for augmented resolution. The proof of this is modeled on a proof that satisfiability of parity clauses is in $P$:

**Lemma 6.7** *Let $C$ be a theory consisting entirely of parity clauses. Then determining whether or not $C$ is satisfiable is in $P$.*

As discussed in the introduction, the proof is basically a Gaussian reduction argument.

**Definition 6.8** *Let $S$ be a subset of a set of $n$ variables. We will say that a permutation $\omega$ flips a variable $v$ if $\omega(v) = \neg v$, and will denote by $F_S$ that subset of $W_n$ consisting of all permutations that leave the variable order unchanged and flip an even number of variables in $S$.*

**Lemma 6.9** $F_S \leq W_n$.

We now have the following:

**Lemma 6.10** *Let $S = \{x_1, \ldots, x_k\}$ be a subset of a set of $n$ variables. Then the parity clause*
$$\sum x_i \equiv 1$$
*is equivalent to the augmented clause*
$$(x_1 \vee \cdots \vee x_k, F_S)$$
*The parity clause*
$$\sum x_i \equiv 0$$
*is equivalent to the augmented clause*
$$(\neg x_1 \vee x_2 \vee \cdots \vee x_k, F_S)$$

We can finally show:

**Proposition 6.11** *Let $C$ be a theory consisting entirely of parity clauses. Then determining whether or not $C$ is satisfiable is in $P$ for augmented resolution.*

We note in passing that the construction in this section fails in the case of modularity clauses with a base other than 2. One of the (many) problems is that the set of permutations that flip a set of variables of size congruent to $m$ (mod $n$) is not a group unless $m = 0$ and $n < 3$. We need $m = 0$ for the identity to be included, and since both

$$(x_1, \neg x_1) \cdots (x_n, \neg x_n)$$

and

$$(x_2, \neg x_2) \cdots (x_{n+1}, \neg x_{n+1})$$

are included, it follows that

$$(x_1, \neg x_1)(x_{n+1}, \neg x_{n+1})$$

must be included, so that $n = 1$ or $n = 2$.

It is not clear whether this is coincidence, or whether there is a deep connection between the fact that mod 2 clauses can be expressed compactly using augmented clauses and are also solvable in polynomial time.

## 7. Theoretical and Procedural Description

In addition to resolution, an examination of Procedures 2.8 and 2.6 shows that we need to be able to eliminate nogoods when they are irrelevant and to identify instances of augmented clauses that are unit. Let us now discuss each of these issues.

The problems around irrelevance are easier to deal with. In the ground case, we remove clauses when they are no longer relevant; in the augmented version, we remove clauses that no longer possess relevant instances. We will defer until the final paper in this series discussion of a procedure for determining whether $(c, G)$ has a relevant instance.

We will also defer discussion of a specific procedure for computing `unit-propagate`$(P)$, but do include a few theoretical comments at this point. In unit propagation, we have a partial assignment $P$ and need to determine which instances of axioms in $C$ are unit. To do this, suppose that we denote by $S(P)$ the set of <u>S</u>atisfied literals in the theory, and by $U(P)$ the set of <u>U</u>nvalued literals. Now for a particular augmented clause $(c, G)$, we are looking for those $g \in G$ such that $c^g \cap S(P) = \emptyset$ and $|c^g \cap U(P)| \leq 1$. The first condition says that $c^g$ has no satisfied literals; the second, that it has at most one unvalued literal.

**Procedure 7.1 (Unit propagation)** *To compute* UNIT-PROPAGATE$(C, P)$ *for a set $C$ of augmented clauses and an annotated partial assignment $P = \langle (l_1, c_1), \ldots, (l_n, c_n) \rangle$:*

1  **while** there is a $(c, G) \in C$ and $g \in G$ with $c^g \cap S(P) = \emptyset$ and $|c^g \cap U(P)| \leq 1$
2      **do if** $c^g \cap U(P) = \emptyset$
3          **then** $l_i \leftarrow$ the literal in $c^g$ with the highest index in $P$
4              **return** $\langle \texttt{true}, \texttt{resolve}((c, G), c_i) \rangle$
5          **else** $l \leftarrow$ the literal in $c^g$ unassigned by $P$
6              add $(l, (c^g, G))$ to $P$
7  **return** $\langle \texttt{false}, P \rangle$

Dixon, Ginsberg, Luks & Parkes

Note that the addition made to $P$ when adding a new literal includes both $c^g$, the instance of the clause that led to the propagation, and the augmenting group as usual. We can use $(c^g, G)$ as the augmented clause by virtue of Proposition 4.7.

Finally, the augmented version of Procedure 2.8 is:

**Procedure 7.2 (Relevance-bounded reasoning, RBL)** *Let $C$ be a SAT problem, and $D$ a set of learned nogoods. Let $P$ be an annotated partial assignment, and $k$ a fixed relevance bound. To compute RBL$(C, D, P)$:*

```
1   ⟨x, y⟩ ← Unit-Propagate(C ∪ D, P)
2   if x = true
3       then (c, G) ← y
4           if c is empty
5               then return FAILURE
6               else  remove successive elements from P so that c is unit
7                     D ← D ∪ {c}
8                     remove from D all augmented clauses without k-relevant instances
9                     return RBL(C, D, P)
10      else  P ← y
11          if P is a solution to C
12              then return P
13              else  l ← a literal not assigned a value by P
14                    return RBL(C, ⟨P, (l, true)⟩)
```

Examining these two procedures, we see that we need to provide implementations of the following:

1. A routine that computes the group of stable extensions appearing in the definition of augmented resolution, needed by line 4 in the unit propagation procedure 7.1.

2. A routine that finds instances of $(c, G)$ for which $c^g \cap S = \emptyset$ and $|c^g \cap U| \leq 1$ for disjoint $S$ and $U$, needed by line 1 in the unit propagation procedure 7.1.

3. A routine that determines whether $(c, G)$ has an instance for which $\texttt{poss}(c^g, P) \leq k$ for some fixed $k$, as needed by line 8 of Procedure 7.2.

All of these problems are known to be NP-complete, although we remind the reader that we continue to measure complexity in terms of the size of the domain and the number of generators of any particular group; the number of generators is logarithmic in the number of instances of any particular augmented clause. It is also the case that the practical complexity of solving these problems appears to be low-order polynomial.

Our focus in the final paper in this series will be on the development of efficient procedures that achieve the above goals, their incorporation into a zCHAFF-like prover, and an evaluation of the performance of the resulting system.

## 8. Conclusion

Our aim in this paper has been to give a theoretical description of a generalized representation scheme for satisfiability problems. The basic building block of the approach is an "augmented clause," a pair $(c, G)$ consisting of a ground clause $c$ and a group $G$ of permutations on the literals in the theory; the intention is that the augmented clause is equivalent to the conjunction of the results of operating on $c$ with elements of $G$. We argued that the structure present in the requirement that $G$ be a group provides a generalization of a wide range of existing notions, from quantification over finite domains to parity clauses.

We went on to show that resolution could be extended to deal with augmented clauses, and gave a generalization of relevance-bounded learning in this setting (Procedures 7.1 and 7.2). We also showed that the resulting proof system generalized first-order techniques when applied to finite domains of quantification, and could produce polynomially sized proofs of the pigeonhole problem, clique coloring problems, Tseitin's graph coloring problems, and parity clauses in general.

Finally, we described the specific group-theoretic problems that need to be addressed in implementing our ideas. As discussed in the previous section, they are:

1. Implementing the group operation associated with the generalization of resolution,

2. Finding unit instances of an augmented clause, and

3. Determining whether an augmented clause has relevant instances.

We will return to these issues in the final paper in this series (Dixon et al., 2004a), which describes an implementation of our ideas and its computational performance.

## Acknowledgments

We would like to thank the members of CIRL, the technical staff of On Time Systems, Paul Beame of the University of Washington, and David Hofer of the CIS department at the University of Oregon for their assistance with the ideas in this series of papers. We also thank the anonymous reviewers for their comments, which improved the exposition greatly.

This work was sponsored in part by grants from Air Force Office of Scientific Research (AFOSR) number F49620-92-J-0384, the Air Force Research Laboratory (AFRL) number F30602-97-0294, Small Business Technology Transfer Research, Advanced Technology Institute (STTR-ATI) number 20000766, Office of Naval Research (ONR) number N00014-00-C-0233, the National Science Foundation (NSF) under grant number CCR9820945, and the Defense Advanced Research Projects Agency (DARPA) and the Air Force Research Laboratory, Rome, NY, under agreements numbered F30602-95-1-0023, F30602-97-1-0294, F30602-98-2-0181, F30602-00-2-0534, and F33615-02-C-4032. The views expressed are those of the authors.

## Appendix A. Proofs

**Proposition 2.7** *Suppose that $C$ is a Boolean satisfiability problem, and $P$ is a sound annotated partial assignment. Then:*

1. If unit-propagate$(P) = \langle \text{false}, P' \rangle$, then $P'$ is a sound extension of $P$, and

2. If unit-propagate$(P) = \langle \text{true}, c \rangle$, then $c$ is a nogood for $P$.

**Proof.** In the first case, we need to show that any extension of $P$ in the procedure leaves $P$ a sound partial assignment. In other words, when we add $(l, c)$ to $P$, we must show that:

1. $C \models c$,

2. $l$ appears in $c$, and

3. Every other literal in $c$ is false by virtue of an assignment in $P$.

For (1), note that $c \in C$. For (2), $l$ is explicitly set to a literal in $c$. And for (3), since $c \in \text{poss}_0(C, P)$, every other literal in $c$ must be set false by $P$.

In the second case in the proposition, $C \models c$ because $c$ is the result of resolving a clause in $C$ with some reason $c_i$, which is entailed by $C$ by virtue of the soundness of $P$. To see that $c$ is falsified by $P$, note that the clause in $\text{poss}_{-1}(C, P)$ is surely falsified by $P$, and that every literal in the reason $c_i$ for $l_i$ is also falsified except for $l_i$ itself. It follows that the result of resolving these two clauses will also be falsified by the assignments in $P$. □

**Theorem 2.9** RBL *is sound and complete in that it will always return a solution to a satisfiable theory $C$ and always report failure if $C$ is unsatisfiable.* RBL *also uses an amount of memory polynomial in the size of $C$ (although exponential in the relevance bound $k$).*

**Proof.** Soundness is immediate. For completeness, note that every nogood learned eliminates an additional portion of the search space, and the backtrack is constrained to not go so far that the newly learned nogood is itself removed as irrelevant.

For the last claim, we extend Definition 2.3 somewhat, defining a *reason* for a literal $l$ to be any learned clause involving $l$ where $l$ was the most recently valued literal at the point that the clause was learned. We will now show that for any literal $l$, there are never more than $(2n)^k$ reasons for $l$, where $n$ is the number of variables in the problem.

To see this, let $R$ be the set of reasons for $l$ at some point. Let $r$ be any reason in this set; between the time that $r \in R$ was learned and the current point, at most $k$ literals in $r$ could have been unassigned by the then-current partial assignment. It follows that there is some fixed partial assignment $P'$ that holds throughout the "life" of each $r \in R$ and such that each $r$ has at most $k$ literals unassigned values by $P'$. Let $S$ be the set of literals assigned values by $P'$.

Given a reason $r_i \in R$, we will view $r_i$ simply as the set of literals that it contains, so that $r_i - S$ is the set of literals appearing in $r_i$ but outside of the stable partial assignment $P'$. Now if $r_j$ was learned before $r_i$, some literal $l_i \in r_i - S$ must not be in $r_j - S$; otherwise, $r_j$ together with the stable partial assignment $P'$ would have precluded the set of variable assignments that led to the conclusion $r_i$. In other words, $r_i - S$ is unique for each reason in the set $R$.

But we also know that $|r_i - S| \leq k$, so that each reason corresponds to choosing at most $k$ literals from the complement of $S$. If there are $n$ variables in the problem, there are most $2n$ literals in this set, so that the number of reasons is bounded by $(2n)^k$. It follows that the total number of reasons learned is bounded by $(2n)^{k+1}$, and the conclusion follows. □

**Theorem A.1 (Lagrange)** *If $G$ is a finite group and $S \leq G$, then $|S|$ divides $|G|$.* □

**Proposition 4.3** Let $S$ be a set of ground clauses, and $(c, G)$ an equivalent augmented clause, where $G$ is represented by generators. It is possible in polynomial time to find a set of generators $\{\omega_1, \ldots, \omega_k\}$ where $k \leq \log_2 |G|$ and $G = \langle \omega_1, \ldots, \omega_k \rangle$.

**Proof.** Even the simplest approach suffices. If $G = \langle g_i \rangle$, checking to see if $g_i \in \langle \omega_1, \ldots, \omega_j \rangle$ for each generator $g_i$ can be performed in polynomial time using a well-known method of Sims (Dixon et al., 2004a; Luks, 1993; Seress, 2003); if $g_i$ is already in the generated set we do nothing and otherwise we add it as a new generator. By virtue of Lagrange's theorem, a subgroup can never be larger than half the size of a group that properly contains it, so adding a new generator to the set of $\omega_i$'s always at least doubles the size of the generated set. It follows that the number of generators needed can never exceed $\log_2 |G|$. □

**Proposition 4.7** Let $(c, G)$ be an augmented clause. Then if $c'$ is any instance of $(c, G)$, $(c, G) \equiv (c', G)$.

**Proof.** Since $c'$ is an instance of $(c, G)$, we must have $c' = c^g$ for some $g \in G$. Thus the instances of $(c', G)$ are clauses of the form $c'^{g'} = c^{gg'}$. But $c^{gg'} = c^{g''}$ for $g'' = gg' \in G$. Similarly, an instance of $(c, G)$ is a clause of the form $c^{g'} = c'^{g^{-1}g'} = c'^{g''}$. □

**Definition A.2** Let $G \leq \text{Sym}(S)$. We will say that $G$ acts *transitively* on $S$ if, for any $x, y \in S$, there is a $g \in G$ with $x^g = y$.

**Proposition 4.8** Let $(c, G)$ be an augmented clause with $d$ distinct instances. Then there is a subgroup $H \leq G$ that can be described using $O(\log_2(d))$ generators such that $(c, H) \equiv (c, G)$. Furthermore, given $d$ and generators for $G$, there is a Monte Carlo polynomial-time algorithm for constructing the generators of such an $H$.

**Proof.** The basic ideas in the proof follow methods introduced by Babai, Luks and Seress (1997). The proof of this particular result is a bit more involved than the others in this paper, and following it is likely to require an existing familiarity with group theory.

Let $D$ be the set of instances of $(c, G)$, so that $G$ acts transitively on $D$. Now consider a sequence $g_1, g_2, \ldots$ of uniformly distributed random elements of $G$ and, for each $r \geq 0$, let $H_r = \langle g_1, g_2, \ldots, g_r \rangle$ (in particular, $H_0 = \langle \emptyset \rangle = \mathbf{1}$). Suppose that $H_{r-1}$ does not act transitively on $D$ and let $K$ be any orbit of $H_{r-1}$ in $D$. Since $G_{\{K\}}$ is a proper subgroup of $G$, Lagrange's theorem implies that the probability that $g_r \in G_{\{K\}}$ is $\leq \frac{1}{2}$. Hence, the probability that $H_r$ enlarges this $K$ is $\geq \frac{1}{2}$. On average then, at least $\frac{1}{2}$ of the orbits will be enlarged in passing from $H_{r-1}$ to $H_r$. Since the orbits partition the entire set $D$, an orbit can only be enlarged if it is merged with one or more other orbits. Thus the fact that at least half of the orbits are enlarged implies that the total number of such orbits is reduced by at least $\frac{1}{4}$. Thus for each $r$, the expected number of orbits of $H_r$ in $D$ is $\leq d(3/4)^r$. As a consequence, with high probability, there exists $r = O(\log_2 d)$ such that $H_r$ acts transitively on $D$. (The probability of failure can be kept below $\epsilon$ for any fixed $\epsilon > 0$.)

An implementation of the algorithm implicit in this proof requires the ability to select uniformly distributed random elements of $G$. These are available at a cost $O(v^2)$ per element, given the standard data structures for permutation group computation (Seress, 2003).[16] □

---

16. The reason that we only have a Monte Carlo method is that there is no known deterministic polynomial time test for testing whether $H \leq G$ acts transitively on $D$; note that $D$ may be exponential in the number of variables in the problem.

**Proposition 5.1** *For ground clauses $c_1$ and $c_2$ and a permutation $\omega \in W_n$,*

$$\texttt{resolve}(\omega(c_1), \omega(c_2)) = \omega(\texttt{resolve}(c_1, c_2))$$

**Proof.** Suppose that the literal being resolved on is $l$, so that if we think of $c_1$ and $c_2$ as being represented simply by the literals they contain, the resolvent corresponds to

$$c_1 \cup c_2 - \{l, \neg l\}$$

Permuting with $\omega$ gives us

$$\omega(c_1) \cup \omega(c_2) - \{\omega(l), \omega(\neg l)\} = \omega(c_1) \cup \omega(c_2) - \{\omega(l), \neg \omega(l)\} \tag{41}$$

where the equality is a consequence of the fact that the permutation in question is a member of $W_n$ instead of simply $S_{2n}$. The right hand side of (41) is simply $\texttt{resolve}(\omega(c_1), \omega(c_2))$. □

**Proposition 5.3** *There are augmented clauses $c_1$ and $c_2$ such that the set $S$ of resolvents of instances of the two clauses does not correspond to any single augmented clause $(c, G)$.*

**Proof.** Consider resolving the augmented clause

$$c_1 = (a \vee b, (bc))$$

with the two instances $a \vee b$ and $a \vee c$, with the augmented clause

$$c_2 = (\neg a \vee d, (dc))$$

corresponding to $\neg a \vee d$ and $\neg a \vee c$. The ground clauses that can be obtained by resolving instances of $c_1$ and of $c_2$ are $b \vee d$, $b \vee c$, $c \vee b$, and $c$. Since these clauses are not of uniform length, they are not instances of a single augmented clause. □

**Proposition 5.8** $\texttt{stab}(K_i, G_i) \leq \text{Sym}(L)$.

**Proof.** Suppose we have $\omega_1, \omega_2 \in \texttt{stab}(K_i, G_i)$. Now for some fixed $i$ and $g_1 \in G_i$, $\omega_1|_{K_i^{G_i}} = g_1|_{K_i^{G_i}}$ and similarly for $\omega_2$ and some $g_2$. But now for any $x \in K_i^{G_i}$,

$$\begin{aligned} x^{\omega_1 \omega_2} &= x^{g_1 \omega_2} \\ &= x^{g_1 g_2} \end{aligned}$$

so that $\omega_1 \omega_2 \in \texttt{stab}(K_i, G_i)$. The first equality holds by virtue of the definition of a stable extension, and the second holds because $x^{g_1}$ is necessarily in the $G_i$-closure of $K_i$. Inversion is similar. □

**Definition 5.4** *A definition of augmented resolution will be called* satisfactory *if any resolvent $(c, G)$ of $(c_1, G_1)$ and $(c_2, G_2)$ satisfies the following conditions:*

1. *It is* sound, *in that $(c_1, G_1) \wedge (c_2, G_2) \models (c, G)$.*

---

Note also that the algorithm explicitly requires that we know $d$ in advance. This is necessary since the quantity is not known to be computable in polynomial time. However, there are methods for computing $d$ that seem to be efficient in practice.

2. It is *complete*, in that $\texttt{resolve}(c_1, c_2)$ is an instance of $(c, G)$.

3. It is *monotonic*, in that if $G_1 \leq H_1$ and $G_2 \leq H_2$, then $(c, G)$ is also a resolvent of $(c_1, H_1)$ and $(c_2, H_2)$.

4. It is *polynomial*, in that it is possible to confirm that $(c, G)$ is a resolvent of $(c_1, G_1)$ and $(c_2, G_2)$ in polynomial time.

5. It is *stable*, in that if $c_1^G = c_2^G$, then $(\texttt{resolve}(c_1, c_2), G)$ is a resolvent of $(c_1, G)$ and $(c_2, G)$

6. It is *strong*, in that if no element of $c_1^{G_1}$ is moved by $G_2$ and similarly no element of $c_2^{G_2}$ is moved by $G_1$, then $(\texttt{resolve}(c_1, c_2), \langle G_1, G_2 \rangle)$ is a resolvent of $(c_1, G_1)$ and $(c_2, G_2)$.

**Proposition 5.10** *Definition 5.9 of (noncanonical) resolution is satisfactory. The definition of canonical resolution satisfies all of the conditions of Definition 5.4 except for monotonicity.*

**Proof.** We deal with the conditions of the definition one at a time.

**1. Soundness**  Any instance of $(c, G)$ must be of the form

$$\omega(\texttt{resolve}(c_1, c_2)))$$

for some $\omega$ that simultaneously extends $G_1$ acting on $c_1$ and $G_2$ acting on $c_2$. But by Proposition 5.1, this is just

$$\texttt{resolve}(\omega(c_1), \omega(c_2))$$

The first of these clauses is an instance of $(c_1, G_1)$, and the second is an instance of $(c_2, G_2)$, so the proposition follows from the soundness of resolution.

**2. Completeness**  $\texttt{resolve}(c_1, c_2)$ is an instance of $(c, G)$ because $c = \texttt{resolve}(c_1, c_2)$ and $1 \in G$.

**3. Monotonicity**  If $(c, G)$ is a resolvent of $(c_1, H_1)$ and $(c_2, H_2)$, then $G = \texttt{stab}(c_i, K_i)$ where $K_i \leq H_i$. But since $H_i \leq G_i$, it follows that $K_i \leq G_i$ and $(c, G)$ is a resolvent of $(c_1, G_1)$ and $(c_2, G_2)$ as well.

**4. Polytime checking**  We assume that we are provided with the intermediate groups $H_1$ and $H_2$, so that we must simply check that $G \leq \texttt{stab}(c_i, H_i)$. Since $\texttt{stab}(c_i, H_i)$ is a group by virtue of Proposition 5.8, it suffices to check that each generator of $G$ is in $\texttt{stab}(c_i, H_i)$. But this is straightforward. Given a generator $g$, we need simply check that restricting $g$ to $c_i^{G_i}$, the image of $c_i$ under $G_i$, produces a permutation that is the restriction of an element of $G_i$. As we remarked in proving Proposition 4.3, this test is known to be in P.

**5. Stability**  It is clear that $(\texttt{resolve}(c_1, c_2), G)$ is a resolvent of $(c_1, G)$ and $(c_2, G)$, since every element of $G$ is clearly a stable extension of $(c_1, G)$ and $(c_2, G)$.

We need the additional condition that $c_1^G = c_2^G$ to show that $(\texttt{resolve}(c_1, c_2), G)$ is the *canonical* resolvent; there is no explicit requirement that the group of stable extensions of $(c_1, G)$ and $(c_2, G)$ be a subgroup of $G$. But if $c_1^G = c_2^G$, the stable extensions must agree with elements of $G$ on $c_1^G = c_2^G$, and hence must agree with elements of $G$ on $c_1$ and $c_2$ themselves, and therefore on $\texttt{resolve}(c_1, c_2)$ as well. Thus the canonical resolvent is equivalent to $(\texttt{resolve}(c_1, c_2), G)$.

**6. Strength** It is clear that the group of stable extensions can never be bigger than $\langle G_1, G_2 \rangle$, since every permutation that either agrees with $c_1$ on $G_1$ and with $c_2$ on $G_2$ is contained in this group. But if $g = \prod g_i$ is an element of $\langle G_1, G_2 \rangle$, with $g_i \in G_1$ for $i$ odd and $g_i \in G_2$ for $i$ even, then $c_1^g = c_1^{(\prod_{i \text{ odd}} g_i)} = c_1^{g'}$ for $g' = \prod_{i \text{ odd}} g_i \in G_1$. The $(c_2, G_2)$ case is similar, so $(\texttt{resolve}(c_1, c_2), \langle G_1, G_2 \rangle)$ is the canonical resolvent of $(c_1, G)$ and $(c_2, G)$. □

**Proposition 6.3** *Let $p$ and $q$ be quantified clauses such that there is a term $t_p$ in $p$ and $\neg t_q$ in $q$ where $t_p$ and $t_q$ have common instances. Suppose also that $(p_g, P)$ is an augmented clause equivalent to $p$ and $(q_g, Q)$ is an augmented clause equivalent to $q$, where $p_g$ and $q_g$ resolve. Then if no terms in $p$ and $q$ except for $t_p$ and $t_q$ have common instances , the result of resolving $p$ and $q$ in the conventional lifted sense is equivalent to $\texttt{resolve}((p_g, P), (q_g, Q))$.*

**Proof.** The proof is already contained in the discussion surrounding the example in the main text. The base instance of the augmented resolvent is clearly an instance of the quantified resolution; for the group, we have already remarked that the group of stable extensions of the two embeddings corresponds simply to any bindings for the variables in the resolvents that can be extended to a permutation of all of the variables in question. This means that the bindings must be consistent with regard to the values selected for shared terms, and no two distinct quantified literals are mapped to identical ground atoms. The latter condition follows from the assumption that the non-resolving literals have no common ground instances. □

**Proposition 6.4** *There is an augmented resolution proof of polynomial size of the mutual unsatisfiability of (34) and (35).*

**Proof.** We begin by explaining how the proof goes generally, and only subsequently provide the details. From the fact that the first pigeon has to be in one of the $n$ holes, we can conclude that one of the first two pigeons must be in one of the last $n - 1$ holes (since these first two pigeons can't both be in the first hole). Now one of the first three pigeons must be in one of the last $n - 2$ holes, and so on until we conclude that one of the first $n$ pigeons must be in the last hole. Similarly, one of the first $n$ pigeons must be in each hole, leaving no hole for the final pigeon.

To formalize this, we will write $A_k$ for the fact that one of the first $k$ pigeons must be in one of the last $n + 1 - k$ holes:

$$A_k \equiv \bigvee_{\substack{1 \leq i \leq k \\ k \leq j \leq n}} p_{ij}$$

Our basic strategy for the proof will be to show that if we denote the original axioms (34) and (35) by $PHP$:[17]

1. $PHP \vdash A_1$,

2. $PHP \wedge A_k \vdash A_{k+1}$,

---

17. Our notation here is vaguely similar to that used by Krishnamurthy (1985), although the both problem being solved and the techniques used are different.

3. $PHP \wedge A_n \vdash \bot$, where $\bot$ denotes a contradiction.

In addition, since the same group $G$ appears throughout the original axiomatization, we will drop it from the derivation, but will feel free to resolve against $c^g$ for any $g \in G$ and derived conclusion $c$.

For the first claim, note that $A_1$ is given by

$$\bigvee_{1 \leq j \leq n} p_{1j}$$

which is an instance of (35).

For the second, we have $A_k$, which is

$$\bigvee_{\substack{1 \leq i \leq k \\ k \leq j \leq n}} p_{ij}$$

and we need to remove all of the variables $p_{jk}$ that refer to the $k$th hole. To do this, we resolve the above clause with each of

$$\begin{aligned}
\neg p_{1k} &\vee \neg p_{k+1,k} \\
\neg p_{2k} &\vee \neg p_{k+1,k} \\
&\vdots \\
\neg p_{kk} &\vee \neg p_{k+1,k}
\end{aligned}$$

to get

$$\neg p_{k+1,k} \vee \bigvee_{\substack{1 \leq i \leq k \\ k+1 \leq j \leq n}} p_{ij}$$

Now note that the only holes mentioned in the disjunction on the right of the above expression are the $k+1$st and higher, so that we can apply the group $G$ to conclude

$$\neg p_{k+1,m} \vee \bigvee_{\substack{1 \leq i \leq k \\ k+1 \leq j \leq n}} p_{ij}$$

for any $1 \leq m \leq k$. Now if we resolve each of these with the instance of (35) given by

$$p_{k+1,1} \vee \cdots \vee p_{k+1,n}$$

we get

$$\bigvee_{k+1 \leq j \leq n} p_{k+1,j} \vee \bigvee_{\substack{1 \leq i \leq k \\ k+1 \leq j \leq n}} p_{ij}$$

which is to say

$$\bigvee_{\substack{1 \leq i \leq k+1 \\ k+1 \leq j \leq n}} p_{ij}$$

or $A_{k+1}$.

Finally, we need to derive a contradiction from $A_n$, which is to say

$$\bigvee_{1 \leq i \leq n} p_{in}$$

Resolving with each of

$$\begin{aligned} \neg p_{1n} &\vee \neg p_{n+1,n} \\ \neg p_{2n} &\vee \neg p_{n+1,n} \\ &\vdots \\ \neg p_{nn} &\vee \neg p_{n+1,n} \end{aligned}$$

now gives $\neg p_{n+1,n}$, and we can thus conclude $\neg p_{ij}$ for any $i$ and $j$ by acting with the group $G$. Resolving into any instance of (35) now gives the desired contradiction. □

**Lemma A.3** *Assuming that we only branch on positive literals in unsatisfied clauses, let $p_{jk}$ be any of the first $n-2$ branch decisions in solving the pigeonhole problem. The set of unit propagations that result from this branch decision is exactly the set $S_k = \{\neg p_{ik} | i \neq j\}$.*

**Proof.** We prove this by induction on the number of branch decisions. For the base case, we take $n \geq 3$ and consider the first branch decision $p_{jk}$. For each $\neg p_{ik} \in S_k$ there is an instance of (35) of the form $\neg p_{jk} \vee \neg p_{ik}$ that causes the unit propagation $\neg p_{ik}$. No other instances of (35) contain literals that refer to hole $k$, so (35) produces no further unit propagations. Each instance of (34) has a total of $n$ literals with at most one literal that refers hole $k$. Because $n \geq 3$, each instance must have at least two unvalued literals and therefore does not generate a unit propagation.

For the inductive case, we assume that Lemma A.3 holds for the first $m$ branches with $m < n-2$. Under this assumption, each branch decision $p_{jk}$ and its subsequent unit propagations value exactly the variables involving hole $k$. We can therefore make the same argument as we did in the base case. Let $p_{jk}$ be the $m+1^{st}$ branch decision. Clause (35) produces exactly the set $S_k = \{\neg p_{ik}|i \neq j\}$ via unit propagation, and because $m+1 \leq n-2$, each instance of (34) has at least two unvalued literals and therefore does not generate any unit propagations.

The key observation is that each branch decision and its subsequent unit propagations value all the variables (and only the variables) that refer to a particular hole. □

**Lemma A.4** *Let $P = \{l_1, l_2, \ldots, l_m\}$ be a partial assignment obtained in solving the pigeonhole problem, where every branch decision is on a positive literal in an unsatisfied clause. For every branch decision $l_i$ in $P$, the subproblem below the open branch $\{l_1, l_2, \ldots, l_{i-1}, \neg l_i\}$ can be solved by unit propagation.*

**Proof.** Assume we are about to begin exploring the subproblem below

$$P = \{l_1, l_2, \ldots, l_i, \neg p_{jk}\}$$

for some branch variable $p_{jk}$. The subproblem below $P = \{l_1, l_2, \ldots, l_i, p_{jk}\}$ has already been explored, found to be unsatisfiable, and we've generated a nogood defining the reason for the failure. This nogood will be an augmented clause of the form

$$(a_1 \vee \cdots \vee a_m \vee \neg p_{jk}, G) \tag{42}$$

The $a_i$ are unsatisfied under $P = \{l_1, l_2, \ldots, l_i\}$, and $G$ is the global symmetry group for the problem.

But now recall that by virtue of Lemma A.3, each of our original branch decisions together with its subsequent unit propagations valued all of the variables that referred to one particular hole and no more. Consider the set of all holes referred to by the partial assignment $\{l_1, l_2, \ldots, l_i\}$. We will call this set $H$.

When we branched on $p_{jk}$, pigeon $j$ had not yet been assigned to a hole. This follows from our assumption that we branch only on positive literals in unsatisfied clauses. Thus for all $h \in H$, $\neg p_{jh} \in P$; in other words, pigeon $j$ was excluded from all of the holes in $H$ prior to our decision to place it in hole $k$. The derived nogood (42) asserts that pigeon $j$ cannot go in hole $k$ either.

But as in the small example worked in the main text, the nogood (42) represents more than a single clause. It represents the set of clauses that can be generated by applying permutations in $G$ to $a_1 \vee \cdots \vee a_m \vee \neg p_{jk}$. If we apply a permutation that swaps hole $k$ with any hole $g \notin H$, the literals $a_1, \ldots, a_m$ will be unchanged and will remain unsatisfied under $P = \{l_1, l_2, \ldots, l_i\}$. So the clause

$$a_1 \vee \cdots \vee a_m \vee \neg p_{jg} \tag{43}$$

is also an instance of (42) for any $g \notin H$, and (43) is also unit under the partial assignment $P$. The nogood (42) thus generates a series of unit propagations indicating that pigeon $j$ cannot be in any hole not in $H$. Since the holes in $H$ are already known to be excluded, there is no hole available for pigeon $j$. A contradiction occurs and the subproblem below $P = \{l_1, l_2, \ldots, l_i, \neg p_{jk}\}$ is closed. □

**Proposition 6.5** *Any implementation of Procedure 2.8 that branches on positive literals in unsatisfied clauses on line 12 will produce a proof of polynomial size of the mutual unsatisfiability of (34) and (35), independent of specific branching choices made.*

**Proof.** Note first that any RBL search tree has size polynomial in the number of branch decisions, since the number of variable assignments that can result from unit propagation is bounded by the number of variables. To show that the search tree has size polynomial in the number of pigeons $n$, it thus suffices to show that the number of branch decisions is polynomial in $n$. We will show that under the given branch heuristic, the number of branch decisions is $n - 1$ specifically.

To do this, we first descend into the tree through branching and propagation until a contradiction is reached, and describe the partial assignment that is created and show how a contradiction is drawn. We then show the backtracking process, proving that the empty clause can be derived in a single backtrack. More specifically, we show that every open branch of the search tree can be closed through propagation alone. No further branch decisions are needed.

Lemma A.3 deals with the first $n-2$ branch decisions. What about the $n-1^{st}$ decision? If this branch decision is $p_{jk}$, we again generate the set of unit propagations $S_k = \{\neg p_{ik} | i \neq j\}$. This time we will generate some additional unit propagations. Since we have assigned $n-1$ of the pigeons each to a unique hole, there is only one empty hole remaining. If this is hole $h$, the two remaining pigeons (say pigeons $a$ and $b$) are both forced into hole $h$, while only one can occupy it. This leads to the expected contradiction. But now Lemma A.4 shows that no further branches are necessary, so that the total number of branches is $n-1$, and the RBL search tree is polynomially sized. □

**Proposition 6.6** *There is an augmented resolution proof of polynomial size of the unsatisfiability of (36)–(40).*

**Proof.** The proof proceeds similarly to the proof of Proposition 6.4, although the details are far more intricate. As before, we will work with ground axioms only and will suppress the augmentation by the global symmetry group.

The analog to $p_{ij}$ is that the $i$th node of the clique gets the $j$th color, or

$$(q_{i1} \wedge c_{1j}) \vee \cdots \vee (q_{im} \wedge c_{mj})$$

which we will manipulate in this form although it's clearly not CNF.

Now $A_1$, the statement that the first pigeon is in some hole, or that the first node of the clique gets some color, is

$$[q_{11} \wedge (c_{11} \vee \cdots \vee c_{1n})] \vee \cdots \vee [q_{1m} \wedge (c_{m1} \vee \cdots \vee c_{mn})]$$

The expression for $A_k$, which was

$$\bigvee_{\substack{1 \leq i \leq k \\ k \leq j \leq n}} p_{ij} \tag{44}$$

similarly becomes

$$[(q_{11} \vee \cdots \vee q_{k1}) \wedge (c_{1k} \vee \cdots \vee c_{1n})] \vee \cdots \vee [(q_{1m} \vee \cdots \vee q_{km}) \wedge (c_{mk} \vee \cdots \vee c_{mn})] \tag{45}$$

saying that for some $i$ and $j$ as in (44), there is an index $h$ such that $q_{ih} \wedge c_{hj}$; $h$ is the index of the graph node to which a clique element of a suitable color gets mapped.

In order to work with the expressions (45), we do need to convert them into CNF. Distributing the $\wedge$ and $\vee$ in (45) will produce a list of conjuncts, each of the form

$$B_1 \vee \cdots \vee B_m \tag{46}$$

where each $B_i$ is of the form either $q_{1i} \vee \cdots \vee q_{ki}$ or $c_{ik} \vee \cdots \vee c_{in}$. There are $2^m$ possible expressions of the form (46).

Each of these $2^m$ expressions, however, is an instance of the result of acting with the global group $G$ on one of the following:

$$\begin{matrix}
(c_{1k} \vee \cdots \vee c_{1n}) & \vee & (c_{2k} \vee \cdots \vee c_{2n}) & \vee & \cdots & \vee & (c_{mk} \vee \cdots \vee c_{mn}) \\
(q_{11} \vee \cdots \vee q_{k1}) & \vee & (c_{2k} \vee \cdots \vee c_{2n}) & \vee & \cdots & \vee & (c_{mk} \vee \cdots \vee c_{mn}) \\
& & & \vdots & & & \\
(q_{11} \vee \cdots \vee q_{k1}) & \vee & (q_{12} \vee \cdots \vee q_{k2}) & \vee & \cdots & \vee & (q_{1m} \vee \cdots \vee q_{km})
\end{matrix} \tag{47}$$

We will view these as all instances of a general construct indexed by $h$, with $h$ giving the number of initial clauses based on the $q$'s instead of the $c$'s. So the first row corresponds to $h = 0$, the second to $h = 1$ and so on, with the last row corresponding to $h = m$. It follows from this that $A_k$ is effectively

$$\bigwedge_{0 \leq h \leq m} \left( \bigvee_{\substack{1 \leq i \leq k \\ 1 \leq j \leq h}} q_{ij} \vee \bigvee_{\substack{h+1 \leq i \leq m \\ k \leq j \leq n}} c_{ij} \right) \tag{48}$$

It is important to realize that we haven't actually started to prove anything yet; we're just setting up the machinery needed to duplicate the proof of Proposition 6.4. The only remaining piece is the analog in this framework of the axiom $\neg p_{ij} \vee \neg p_{kj}$, saying that each hole can only contain one pigeon. That is

$$\neg q_{ih} \vee \neg c_{hj} \vee \neg q_{kg} \vee \neg c_{gj} \tag{49}$$

saying that if node $i$ of the clique is mapped to node $h$ of the graph, and $k$ is mapped to $g$, then $g$ and $h$ cannot both get the same color.

If $g \neq h$, we can derive (49) by resolving (36) and (40). If $g = h$, then (49) becomes $\neg q_{ih} \vee \neg c_{hj} \vee \neg q_{kh}$ and is clearly a weakening of (39). Thus $\neg p_{ij} \vee \neg p_{kj}$ becomes the pair of clauses

$$\neg q_{ih} \vee \neg c_{hj} \vee \neg q_{kg} \vee \neg c_{gj}$$
$$\neg q_{ih} \vee \neg q_{kh}$$

both of which can be derived in polynomial time and are, as usual, acted on by the group $G$. We are finally ready to proceed with the main proof.

For the base step, we must derive $A_1$, or the conjunction of

$$\bigvee_{\substack{1 \leq i \leq 1 \\ 1 \leq j \leq h}} q_{ij} \vee \bigvee_{\substack{h+1 \leq i \leq m \\ 1 \leq j \leq n}} c_{ij}$$

which is to say

$$\begin{array}{cccccc}
(c_{11} \vee \cdots \vee c_{1n}) & \vee & (c_{21} \vee \cdots \vee c_{2n}) & \vee & \cdots \vee & (c_{m1} \vee \cdots \vee c_{mn}) \\
q_{11} & \vee & (c_{21} \vee \cdots \vee c_{2n}) & \vee & \cdots \vee & (c_{m1} \vee \cdots \vee c_{mn}) \\
& & & \vdots & & \\
q_{11} & \vee & q_{12} & \vee & \cdots \vee & q_{1m}
\end{array}$$

Except for the last row, each of these is obviously a weakening of (37) saying that every node in the graph gets a color. The final row is equivalent to (38) saying that each element of the clique gets a node in the graph.

For the inductive step, we must show that $A_k \vdash A_{k+1}$. Some simplifying notation will help, so we introduce

$$C_{ij..k} \equiv (c_{ij} \vee \cdots \vee c_{ik})$$

and

$$Q_{i..jk} \equiv (q_{ik} \vee \cdots \vee q_{jk})$$

Now $A_k$ as in (47) is actually

$$
\begin{array}{cccccc}
C_{1k..n} & \vee & C_{2k..n} & \vee & \cdots & \vee & C_{mk..n} \\
Q_{1..k1} & \vee & C_{2k..n} & \vee & \cdots & \vee & C_{mk..n} \\
& & & & \vdots & & \\
Q_{1..k1} & \vee & Q_{1..k2} & \vee & \cdots & \vee & Q_{1..km}
\end{array}
\tag{50}
$$

Following the pigeonhole proof, we need to reduce the number of holes (i.e., colors) by one and increase the number of pigeons (i.e., clique elements) by one. For the first step, we need to resolve away the appearances of $c_{ik}$ from (50). We claim that from (50) it is possible to derive

$$
\begin{array}{cccccccccc}
C_{1,k+1..n} & \vee & C_{2,k+1..n} & \vee & \cdots & \vee & C_{m,k+1..n} & \vee & \neg q_{k+1,m} & \vee & \neg c_{mk} \\
Q_{1..k1} & \vee & C_{2,k+1..n} & \vee & \cdots & \vee & C_{m,k+1..n} & \vee & \neg q_{k+1,m} & \vee & \neg c_{mk} \\
& & & & \vdots & & & & & & \\
Q_{1..k1} & \vee & Q_{1..k2} & \vee & \cdots & \vee & Q_{1..km} & \vee & \neg q_{k+1,m} & \vee & \neg c_{mk}
\end{array}
\tag{51}
$$

We show this by working from the bottom of the arrays ($h = m$ in the description (48)) to the top ($h = 0$). For the last row, the expression in (51) is clearly a weakening of the final row in (50).

Suppose that we have done $h = i$ and are now considering $h = i - 1$ corresponding to the derivation of the disjunction

$$Q_{1..k1} \vee \cdots \vee Q_{1..k,i-1} \vee C_{i,k+1..n} \vee \cdots \vee C_{m,k+1..n} \vee \neg q_{k+1,m} \vee \neg c_{mk}$$

from

$$Q_{1..k1} \vee \cdots \vee Q_{1..k,i-1} \vee C_{ik..n} \vee \cdots \vee C_{mk..n} \tag{52}$$

Now recall that $C_{jk..n}$ is

$$c_{jk} \vee \cdots \vee c_{jn}$$

and we want to remove the $c_{jk}$ term by resolving with (49), adding literals of the form $\neg q_{k+1,m} \vee \neg c_{mk}$. The instance of (49) that we use is

$$\neg q_{k+1,m} \vee \neg c_{mk} \vee \neg q_{lj} \vee \neg c_{jk}$$

and the resolvent is

$$C_{j,k+1..n} \vee \neg q_{k+1,m} \vee \neg c_{mk} \vee \neg q_{lj}$$

The result of resolving into (52) is therefore

$$Q_{1..k1} \vee \cdots \vee Q_{1..k,i-1} \vee C_{ik..n} \vee \cdots \vee C_{j-1,k..n} \vee C_{j,k+1..n} \vee C_{j+1,k..n} \vee \cdots \vee C_{mk..n} \vee \neg q_{k+1,m} \vee \neg c_{mk} \vee \neg q_{lj} \tag{53}$$

for any $l$ and $i \leq j \leq m$.

Now we have already done $h = i$, which means that we have derived

$$Q_{1..k1} \vee \cdots \vee Q_{1..ki} \vee C_{i+1,k+1..n} \vee \cdots \vee C_{m,k+1..n} \vee \neg q_{k+1,m} \vee \neg c_{mk}$$

Operating on this with the usual group allows us to exchange the $q$ and $c$ variables arbitrarily, so it matters not that the first $i$ terms involve the $q$ variables and the last $m - i$ involve the $c$ variables, but only that the *number* of terms involving $q$ and $c$ variables are $i$ and $m - i$ respectively. Thus we have also derived

$$Q_{1..k1} \vee \cdots \vee Q_{1..k,i-1} \vee C_{i,k+1..n} \vee \cdots \vee C_{j-1,k+1..n} \vee Q_{1..kj} \vee C_{j+1,k+1..n} \vee \cdots \vee C_{m,k+1..n} \vee \neg q_{k+1,m} \vee \neg c_{mk} \tag{54}$$

(where we have essentially swapped node $i$ and node $j$ in (53)). By taking $l = 1, \ldots, k$ in (53), we can resolve (54) with (53) to eliminate both the trailing $\neg q_{lj}$ in (53) and the $Q_{1..kj}$ term above. Since the literals in $C_{h,k+1..n}$ are a subset of those in $C_{hk..n}$, we get

$$Q_{1..k1} \vee \cdots \vee Q_{1..k,i-1} \vee C_{ik..n} \vee \cdots \vee C_{j-1,k..n} \vee C_{j,k+1..n} \vee C_{j+1,k..n} \vee \cdots \vee C_{mk..n} \vee \neg q_{k+1,m} \vee \neg c_{mk}$$

We can continue in this fashion, gradually raising the second index of each $C$ term from $k$ to $k+1$, to finally obtain

$$Q_{1..k1} \vee \cdots \vee Q_{1..k,i-1} \vee C_{i,k+1..n} \vee \cdots \vee C_{m,k+1..n} \vee \neg q_{k+1,m} \vee \neg c_{mk} \tag{55}$$

as in (51).

The hard part is now done; we have exploited the symmetry over the nodes and it remains to use the symmetry over the colors. In the derivation of (55), the color $k$ in the final $\neg c_{mk}$ is obviously irrelevant provided that it is chosen from the set $1, \ldots, k$; higher numbered colors (but only higher numbered colors) already appear in the $C_{j,k+1..n}$. So we have actually derived

$$
\begin{aligned}
Q_{1..k1} &\vee \cdots \vee Q_{1..k,i-1} \vee C_{i,k+1..n} \vee \cdots \vee C_{m,k+1..n} \vee \neg q_{k+1,m} \vee \neg c_{mk} \\
&\vdots \\
Q_{1..k1} &\vee \cdots \vee Q_{1..k,i-1} \vee C_{i,k+1..n} \vee \cdots \vee C_{m,k+1..n} \vee \neg q_{k+1,m} \vee \neg c_{m2} \\
Q_{1..k1} &\vee \cdots \vee Q_{1..k,i-1} \vee C_{i,k+1..n} \vee \cdots \vee C_{m,k+1..n} \vee \neg q_{k+1,m} \vee \neg c_{m1}
\end{aligned}
$$

and when we resolve all of these with the domain axiom (37)

$$c_{m1} \vee c_{m2} \vee \cdots \vee c_{mn}$$

we get

$$Q_{1..k1} \vee \cdots \vee Q_{1..k,i-1} \vee C_{i,k+1..n} \vee \cdots \vee C_{m,k+1..n} \vee \neg q_{k+1,m} \vee c_{m,k+1} \vee \cdots \vee c_{mn}$$

which is to say

$$Q_{1..k1} \vee \cdots \vee Q_{1..k,i-1} \vee C_{i,k+1..n} \vee \cdots \vee C_{m,k+1..n} \vee \neg q_{k+1,m} \vee C_{m,k+1..n}$$

or

$$Q_{1..k1} \vee \cdots \vee Q_{1..k,i-1} \vee C_{i,k+1..n} \vee \cdots \vee C_{m,k+1..n} \vee \neg q_{k+1,m}$$

Now the $m$ subscript in the final $q$ variable is also of no import, provided that it remains at least $i$. We can therefore resolve

$$
\begin{aligned}
Q_{1..k1} &\vee \cdots \vee Q_{1..k,i-1} \vee C_{i,k+1..n} \vee \cdots \vee C_{m,k+1..n} \vee \neg q_{k+1,i} \\
&\vdots \\
Q_{1..k1} &\vee \cdots \vee Q_{1..k,i-1} \vee C_{i,k+1..n} \vee \cdots \vee C_{m,k+1..n} \vee \neg q_{k+1,m-1} \\
Q_{1..k1} &\vee \cdots \vee Q_{1..k,i-1} \vee C_{i,k+1..n} \vee \cdots \vee C_{m,k+1..n} \vee \neg q_{k+1,m}
\end{aligned}
$$

with the domain axiom (38)
$$q_{k+1,1} \vee \cdots \vee q_{k+1,m}$$
to get
$$Q_{1..k1} \vee \cdots \vee Q_{1..k,i-1} \vee C_{i,k+1..n} \vee \cdots \vee C_{m,k+1..n} \vee q_{k+1,1} \vee \cdots \vee q_{k+1,i-1}$$
which is to say
$$Q_{1..k+1,1} \vee \cdots \vee Q_{1..k+1,i-1} \vee C_{i,k+1..n} \vee \cdots \vee C_{m,k+1..n}$$
This is $A_{k+1}$ as desired.

It remains to show that we can derive a contradiction from $A_n$. If we continue with the above procedure in an attempt to "derive" $A_{n+1}$, when we derive an instance of (55), the $C$ terms will simply vanish because $k+1 > n$. We end up concluding
$$Q_{1..k1} \vee \cdots \vee Q_{1..k,i-1} \vee \neg q_{k+1,m} \vee \neg c_{mk}$$
and the $i = 0$ instance is simply
$$\neg q_{k+1,m} \vee \neg c_{mk}$$

All of the indices here are subject to the usual symmetry, so we know
$$\begin{aligned} \neg q_{ji} &\vee \neg c_{i1} \\ \neg q_{ji} &\vee \neg c_{i2} \\ &\vdots \\ \neg q_{ji} &\vee \neg c_{in} \end{aligned}$$

which we resolve with $c_{i1} \vee \cdots \vee c_{in}$ to get $\neg q_{ji}$. We can resolve instances of this with $q_{m1} \vee \cdots \vee q_{mn}$ to finally get the desired contradiction. □

**Lemma 6.7** *Let $C$ be a theory consisting entirely of parity clauses. Then determining whether or not $C$ is satisfiable is in $P$.*

**Proof.** The proof is essentially a Gaussian reduction argument, and proceeds by induction on $n$, the number of variables in $C$. If $n = 0$, the result is immediate. So suppose that $C$ contains $n + 1$ variables, and let one clause containing $x_1$ be
$$x_1 + \sum_{x \in S} x \equiv k$$
where $k = 0$ or $k = 1$. This is obviously equivalent to
$$x_1 \equiv \sum_{x \in S} x + k$$
which we can now use to eliminate $x_1$ from every other axiom in $C$ in which it appears. Since the resulting theory can be tested for satisfiability in polynomial time, the result follows. □

**Lemma A.5** *Suppose that we have two axioms given by*

$$x_1 + \sum_{x \in S} x + \sum_{x \in T_1} x \equiv k_1 \tag{56}$$

*and*

$$x_1 + \sum_{x \in S} x + \sum_{x \in T_2} x \equiv k_2 \tag{57}$$

*where the sets $S$, $T_1$ and $T_2$ are all disjoint. Then it follows that*

$$\sum_{x \in T_1} x + \sum_{x \in T_2} x \equiv k_1 + k_2 \tag{58}$$

*and furthermore, any proof system that can derive this in polynomial time can also determine the satisfiability of sets of parity clauses in polynomial time.*

**Proof.** Adding (56) and (57) produces (58). That this is sufficient to solve sets of parity clauses in polynomial time is shown in the proof of Lemma 6.7. □

**Lemma 6.9** $F_S \leq W_n$.

**Proof.** $F_S$ is closed under inversion, since every element in $F_S$ is its own inverse. To see that it is closed under composition as well, suppose that $f_1$ flips the variables in a set $S_1$ and $f_2$ flips the variables in a set $S_2$. Then $f_1 f_2$ flips the variables in $S_{12} = S_1 \cup S_2 - (S_1 \cap S_2)$. But now

$$\begin{aligned}
|S_{12}| &= |S_1 \cup S_2 - (S_1 \cap S_2)| \\
&= |S_1 \cup S_2| - |S_1 \cap S_2| \\
&= |S_1| + |S_2| - |S_1 \cap S_2| - |S_1 \cap S_2| \\
&= |S_1| + |S_2| - 2 \cdot |S_1 \cap S_2|
\end{aligned}$$

and is therefore even, so $f_1 f_2 \in F_S$. □

**Lemma 6.10** *Let $S = \{x_1, \ldots, x_k\}$ be a subset of a set of $n$ variables. Then the parity clause*

$$\sum_{i=1}^{k} x_i \equiv 1 \tag{59}$$

*is equivalent to the augmented clause*

$$(x_1 \vee \cdots \vee x_k, F_S) \tag{60}$$

*The parity clause*

$$\sum_{i=1}^{k} x_i \equiv 0$$

*is equivalent to the augmented clause*

$$(\neg x_1 \vee x_2 \vee \cdots \vee x_k, F_S)$$

**Proof.** To see that (59) implies (60), note that (59) certainly implies $x_1 \vee \cdots \vee x_k$. But the result of operating on the disjunction with any element of $F_S$ flips an even number of elements in it, so (60) follows.

For the converse, suppose that (59) fails, so that an even number of the $x_i$'s are true. The disjunction that flips exactly those $x_i$'s that are true will obviously have no satisfied literals, but will have flipped an even number of elements of $S$ so that some instance of the augmented clause (60) is unsatisfied.

The second equivalence clearly follows from the first; replace $x_1$ with $\neg x_1$. □

**Proposition 6.11** *Let $C$ be a theory consisting entirely of parity clauses. Then determining whether or not $C$ is satisfiable is in P for augmented resolution.*

**Proof.** We need to show that the conditions of Lemma A.5 are met. We can assume without loss of generality that $k_1 = 1$ and $k_2 = 0$ in the conditions of the lemma; other cases involve simply flipping the sign of one of the variables involved.

In light of Lemma 6.10, we have the two augmented axioms

$$(x_1 \vee \bigvee_{x \in S} x \vee \bigvee_{x \in T_1} x, F_{x_1 \cup S \cup T_1})$$

and

$$(\neg x_1 \vee \bigvee_{x \in S} x \vee \bigvee_{x \in T_2} x, F_{x_1 \cup S \cup T_2})$$

where $S$, $T_1$, and $T_2$ are all disjoint. The clause obtained in the resolution is clearly

$$\bigvee_{x \in S \cup T_1 \cup T_2} x$$

but what is the group involved?

The elements of the group are the stable extensions of group elements from $F_{x_1 \cup S \cup T_1}$ and $F_{x_1 \cup S \cup T_2}$; in other words, any permutation that leaves the variables unchanged and simultaneously flips an even number of elements of $x_1 \cup S \cup T_1$ and of $x_1 \cup S \cup T_2$. We claim that these are exactly those elements that flip any subset of $S$ and an even number of elements of $T_1 \cup T_2$.

We first show that any stable extension $g$ flips an even number of elements of $T_1 \cup T_2$ (along with some arbitrary subset of $S$). If $g$ flips an odd number of elements of $T_1 \cup T_2$, then it must flip an odd number of elements of one (say $T_1$) and an even number of elements of the other. Now if the parity of the number of flipped elements of $x_1 \cup S$ is even, the total number flipped in $x_1 \cup S \cup T_1$ will be odd, so that $g$ does not match an element of $F_{x_1 \cup S \cup T_1}$ and is therefore not an extension. If the parity of the number of flipped elements of $x_1 \cup S$ is odd, the number flipped in $x_1 \cup S \cup T_2$ will be odd.

To see that any $g$ flipping an even number of elements of $T_1 \cup T_2$ corresponds to a stable extension, we note simply that by flipping $x_1$ or not, we can ensure that $g$ flips an even number of elements in each relevant subset. Since $g$ flips an even number of elements in $T_1 \cup T_2$, it flips subsets of $T_1$ and of $T_2$ that have the same parity. So if it flips an even number of elements of $S \cup T_1$, it also flips an even number of elements of $S \cup T_2$ and we leave $x_1$ unflipped. If $g$ flips an odd number of elements of $S \cup T_1$ and of $S \cup T_2$, we flip $x_1$. Either way, there are corresponding elements of $F_{x_1 \cup S \cup T_1}$ and $F_{x_1 \cup S \cup T_2}$.

Now suppose that we denote by $K_S$ the group that flips an arbitrary subset of $S$. We have shown that the result of the resolution is

$$\left( \bigvee_{x \in S \cup T_1 \cup T_2} x, K_S \times F_{T_1 \cup T_2} \right) \tag{61}$$

Note that the resolution step itself is polytime, since we have given the result explicitly in (61).[18]

Next, we claim that (61) implies

$$\left( \bigvee_{x \in T_1 \cup T_2} x, K_S \times F_{T_1 \cup T_2} \right) \tag{62}$$

where we have removed the elements of $S$ from the disjunction.

We prove this by induction on the size of $S$. If $S = \emptyset$, the result is immediate. Otherwise, if $a \in S$ for some specific $a$, two instances of (61) are

$$\left( a \vee \bigvee_{x \in S - \{a\} \cup T_1 \cup T_2} x, K_S \times F_{T_1 \cup T_2} \right)$$

and

$$\left( \neg a \vee \bigvee_{x \in S - \{a\} \cup T_1 \cup T_2} x, K_S \times F_{T_1 \cup T_2} \right)$$

which we can resolve using the stability property (5) of Definition 5.4 to conclude

$$\left( \bigvee_{x \in S - \{a\} \cup T_1 \cup T_2} x, K_S \times F_{T_1 \cup T_2} \right)$$

so that (62) now follows by the inductive hypothesis.

At this point, however, note that the variables in $S$ do not appear in the clause in (62), so that we can drop $K_S$ from the group in the conclusion without affecting it in any way. Thus we have concluded

$$\left( \bigvee_{x \in T_1 \cup T_2} x, F_{T_1 \cup T_2} \right) \tag{63}$$

Applying Lemma 6.10 once again, we see that (63) is equivalent to

$$\sum_{x \in T_1} x + \sum_{x \in T_2} x \equiv 1$$

as needed by Lemma A.5. The proof is complete. □

---

18. In general, augmented resolution is not known to be polynomial in the number of generators of the groups in question. But it is polynomial for groups of restricted form being considered here.